\documentclass{article}

\usepackage{microtype}
\usepackage{graphicx}
\usepackage{subcaption}
\usepackage{booktabs} % for professional tables

\usepackage{hyperref}

\usepackage[preprint]{icml2026}

\usepackage{amsmath}
\usepackage{amssymb}
\usepackage{mathtools}
\usepackage{amsthm}
\usepackage{setspace}

\usepackage[capitalize,noabbrev]{cleveref}

\theoremstyle{plain}
\newtheorem{theorem}{Theorem}[section]
\newtheorem{proposition}[theorem]{Proposition}

\newtheorem{corollary}[theorem]{Corollary}
\theoremstyle{definition}
\newtheorem{definition}[theorem]{Definition}

\theoremstyle{remark}

\newcommand{\BPL}{\mathbf{L}} %{\mathcal{L}^L}

\newcommand\norm[1]{\left\lVert#1\right\rVert}

\newcommand{\gpool}{\operatorname{pool}} %{\mathcal{L}^L}
\newcommand{\flatten}{\operatorname{flatten}} %{\mathcal{L}^L}
\newcommand{\pool}{\operatorname{pool}} %{\mathcal{L}^L}

\usepackage[disable,textsize=tiny]{todonotes}
\icmltitlerunning{Can Local Learning Match Self-Supervised Backpropagation?}

\begin{document}

\twocolumn[
%  \icmltitle{When can local learning implement exactly self-supervised backpropagation?}
\icmltitle{Can Local Learning Match Self-Supervised Backpropagation?}

  %\icmltitle{Self-supervised backprop implemented with layerwise local learning rules}
  
  % It is OKAY to include author information, even for blind submissions: the
  % style file will automatically remove it for you unless you've provided
  % the [accepted] option to the icml2026 package.

  % List of affiliations: The first argument should be a (short) identifier you
  % will use later to specify author affiliations Academic affiliations
  % should list Department, University, City, Region, Country Industry
  % affiliations should list Company, City, Region, Country

  % You can specify symbols, otherwise they are numbered in order. Ideally, you
  % should not use this facility. Affiliations will be numbered in order of
  % appearance and this is the preferred way.
  \icmlsetsymbol{equal}{*}

  \begin{icmlauthorlist}
    \icmlauthor{Wu S. Zihan}{yyy}
    \icmlauthor{Ariane Delrocq}{yyy}
    \icmlauthor{Wulfram Gerstner}{yyy}
    \icmlauthor{Guillaume Bellec}{sch}
    % \icmlauthor{Firstname5 Lastname5}{yyy}
    % \icmlauthor{Firstname6 Lastname6}{sch,yyy,comp}
    % \icmlauthor{Firstname7 Lastname7}{comp}
    % \icmlauthor{}{sch}
    % \icmlauthor{Firstname8 Lastname8}{sch}
    % \icmlauthor{Firstname8 Lastname8}{yyy,comp}
    %\icmlauthor{}{sch}
    %\icmlauthor{}{sch}
  \end{icmlauthorlist}

  \icmlaffiliation{yyy}{School of Computer and Communication Sciences and School of Life Sciences, EPFL, Lausanne, Switzerland}
  \icmlaffiliation{sch}{Machine Learning Research Unit, TU Wien, Vienna, Austria}

  \icmlcorrespondingauthor{Wu S. Zihan}{zihan.wu@epfl.ch}
  \icmlcorrespondingauthor{Guillaume Bellec}{guillaume.bellec@tuwien.ac.at}

  % You may provide any keywords that you find helpful for describing your
  % paper; these are used to populate the "keywords" metadata in the PDF but
  % will not be shown in the document
  \icmlkeywords{Machine Learning, ICML}

  \vskip 0.3in
]

% this must go after the closing bracket ] following \twocolumn[ ...

% This command actually creates the footnote in the first column listing the
% affiliations and the copyright notice. The command takes one argument, which
% is text to display at the start of the footnote. The \icmlEqualContribution
% command is standard text for equal contribution. Remove it (just {}) if you
% do not need this facility.

% Use ONE of the following lines. DO NOT remove the command.
% If you have no special notice, KEEP empty braces:
\printAffiliationsAndNotice{}  % no special notice (required even if empty)
% {\renewcommand{\thefootnote}{}\footnotetext{\noindent Code is available at \url{https://github.com/zihan-wu/local-SSL}}}

% Or, if applicable, use the standard equal contribution text:
% \printAffiliationsAndNotice{\icmlEqualContribution}

\begin{abstract}
     While end-to-end self-supervised learning with backpropagation (global BP-SSL) has become central for training modern AI systems, theories of {\em local} self-supervised learning (local-SSL) have struggled to build functional representations in deep neural networks.
To establish a link between global and local rules, we first develop a theory for deep linear networks: we identify conditions for local-SSL algorithms (like Forward-forward or CLAPP) to implement exactly the same weight update as a global BP-SSL.
Starting from the theoretical insights, we then 
develop novel variants of local-SSL algorithms
to approximate global BP-SSL in deep non-linear convolutional neural networks.
Variants that improve the similarity between gradient updates of local-SSL with those of global BP-SSL also show better performance on image datasets (CIFAR-10, STL-10, Tiny ImageNet and ImageNet). 
The best local-SSL rule with the CLAPP loss function matches the performance of a comparable global BP-SSL with InfoNCE or CPC-like loss functions, and improves upon state-of-the-art for local-SSL on these benchmarks. 
    %While there are successful local learning rules (LLR) that can theoretically estimate BP in the supervised case, non of them have demonstrated scalable and theoretical so far.
    %While there are successful local learning rules (LLR) that can theoretically estimate BP in the supervised case, non of them have demonstrated scalable and theoretical so far.
    %While there is t. On the other side of the spectrum of biologically plausible algorithms,  plausibility and locality of learning are achieved by optimizing a self-supervised objective function separately for each layer of a deep network. However, the absence of gradient flow between layers makes such approaches not explicitly approximate backpropagation. 
    %In this paper, we study a family of layerwise contrastive learning rules and theoretically analyze under which conditions could these learning rules could approximate the end-to-end backpropagation training. Based on the theoretical analysis, we improve the performance of CLAPP, a recently published layerwise self-supervised learning rule, on various image classification benchmarks, even close to the performance of training a model end-to-end with the same self-supervised objectives. 
\end{abstract}

\section{Introduction}

% First paragraph: High level insights from biology

\begin{figure}[ht]
  \centering
  \includegraphics[width=\columnwidth]{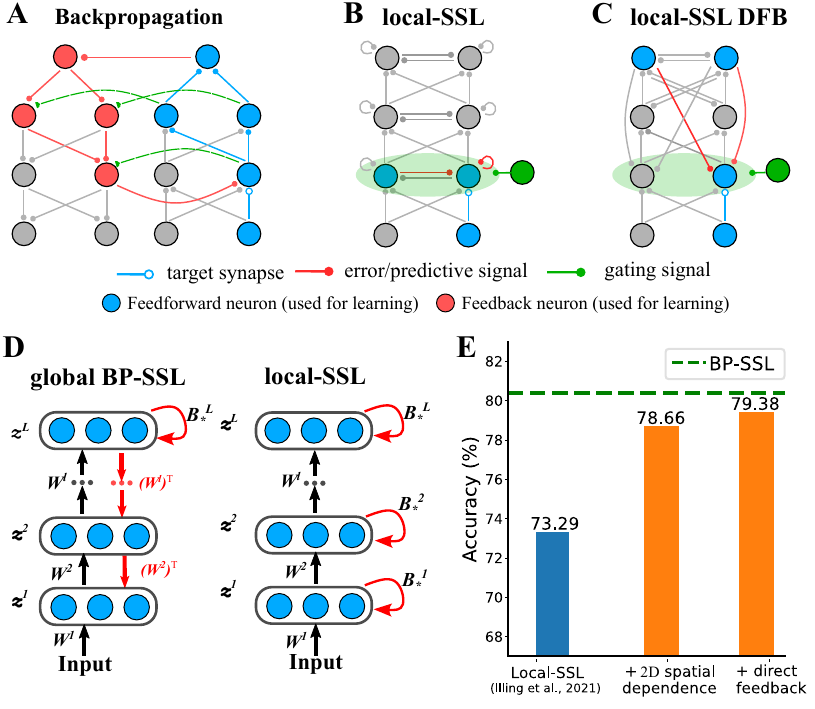}
  \caption{\textbf{Overview of BP and local-SSL} \textbf{(A)} To train the target synapse (blue), BP gradients need to be propagated down a one-to-one matching error network (red) that is gated (green) by the feedforward activations. \textbf{(B)}: In Local-SSL, plasticity is modulated by predictive signals from same-layer lateral projections (red arrow) and global scalar values (green). \textbf{(C)} We show that local-SSL with feedback from the top layer better approximates BP. \textbf{(D) Notations for theoretical analysis}: In Section 3, we compare gradient updates of local-SSL and global BP-SSL. Both algorithms share the same feedforward weight $W^l$ and projection $B_*^L$ in the last layer. Trainable projections $B^l$ are assumed to reach optimum $B_*^l$ instantaneously in our theoretical analysis (Details in Section 3).  \textbf{(E) Performance summary}: On STL-10, our theory-guided improvements (orange bars) improve local-SSL algorithms. More details in Section \ref{sec:empirical_results}.}
  \label{fig:sketch}
\end{figure}

 As a model of brain plasticity, Backpropagation (BP) poses multiple problems~\cite{crick1989bp, lillicrap2020backpropagation,richards2019deep}.
Whereas the forward pass of the BP algorithm might be an acceptable model of rapid biological neural computations \cite{yamins2014performance,schrimpf2018brain}, the backward pass requires a structured feedback machinery that is absent in biological networks (Figure \ref{fig:sketch}A).
Biologically, learning is likely implemented by connectivity changes controlled by a Hebb rule, loosely paraphrased as "neurons that fire together wire together"  \cite{hebb1949organization,Shatz92}. While many classic experiments confirmed the Hebbian paradigm~\cite{lomo1966frequency,castellucci1970neuronal,Bi01,Caporale08,markram2011history}, the modern view emphasizes that Hebbian learning also depends on neuromodulators
\cite{Lisman11,yagishita2014critical,bittner2017behavioral,He15} 
that could signal predictions, expectations, reward, or novelty
\cite{schultz1997neural,gerstner2018eligibility}.
In parallel, in the modeling community, there has been a shift from Hebbian models
\cite{oja1982simplified,barlow1989unsupervised,Miller89,Hyvarinen98}
to more general {\em local} learning rules in which Hebbian plasticity is influenced by neuromodulation or predictive mechanisms \cite{williams1992simple,foldiak1991learning,rao1999predictive,wiskott2002slow,whittington2017predictivecoding,gerstner2018eligibility,Bredenberg21impression,Bredenberg23formalize}.
However, in practice, none of these local algorithms has worked as well as global BP in training deep neural networks~\cite{bartunov2018assessing, lillicrap2020backpropagation, filipovich2022scaling}.

% Local unsupervised, forward-forward
To train deep networks in a {\em supervised} setup with labeled data, local learning algorithms have been proposed~\cite{Roelfsema05,lee2015difference,lillicrap2016fa,Sacramento18,akrout2019wm,meulemans2022minimizing,Song24,Salvatori2024ipc}.
They can approximate BP in theory~\cite{scellier2017equilibrium,whittington2017predictivecoding} and approach its performance on benchmarks~\cite{akrout2019wm,launay2020direct,ernoult2022towards}.
However, for {\em self-supervised learning} without labeled data, the performance gap is larger, and we miss a theory to relate local-SSL algorithms with global representation learning principles.
%the success of this class of learning rules has been limited. 
Existing partially-local learning rules for self-supervised learning cut the backward flow into blocks and gradient propagation is stopped outside of the block~\cite{lowe2019gim,siddiqui2024blockwise,kappel2023block}.
Each network block has one self-supervised InfoNCE loss  as used in Contrastive Predictive Coding (CPC)~\cite{oord2018representation}, SimCLR~\cite{chen2020simclr} or CLIP \cite{radford2021learning}. 
In the extreme case of a ``single layer per block'', this approach yields local self-supervised learning (local-SSL) algorithms, which we see as completely local and BP free.
This construction yields currently the best-performing local unsupervised learning algorithms, and in some cases the weight update could also be interpreted as Hebbian rule variants~\cite{illing_2021,halvagal2023lpl,chen2025self}.
The three major types of local-SSL algorithms are:
 contrastive local and predictive plasticity rules (CLAPP) \cite{illing_2021,Delroq2024critical}, Forward-forward algorithm variants~\cite{hinton2022forward,momeni2023phyll,chen2025self}, and non-contrastive methods like LPL \cite{halvagal2023lpl}.
Variations of such local-SSL algorithms enable representation learning in cost-efficient or energy-efficient physical computing systems~\cite{momeni2023phyll,weilenmann2024single} and digital event-based hardware \cite{graf2024echospike,su2025elfcore}. 

% PhyLL \cite{momeni2023phyll} was inspired by Forward-forward to enable learning in physical networks without BP.
% The physical computing element is interpreted as a non-linear mixing with unknown weights and non-linearity and the full device is therefore a superposition of digital computers and unknown physical layers.
% The locality of local-SSL becomes handy to train the digital computer independently of the unknown physical computation~\cite{momeni2023phyll}.

%% SUMMARY OF RESULTS AND CONTRIBUTION
Building upon existing local-SSL algorithms~\cite{illing_2021,hinton2022forward,halvagal2023lpl,zhu2025stochastic,chen2025self}, we ask two questions: (1) Can we find a theoretical foundation for local-SSL algorithms? (2) Can we reduce the existing performance gap between local-SSL and global BP-SSL? 
%Regarding the lack of theoretical understanding, we discovered that some local-SSL algorithms are theoretically consistent with layer-wise or BP updates in the following sense~(Theorem \ref{th1}): In deep linear networks with orthogonal weights, the weight updates with local-SSL and self-supervised BP are strictly equal.We further analyzed that the performance gap between local-SSL and self-supervised BP can be reduced when local-SSL theoretically or numerically approximates BP.
Our main contributions are: 

\begin{itemize}
    %\item We show that many recently proposed local layerwise learning rules could be phrased in the same framework. The latest version of them performs in a similar range on image classification benchmarks.
    
    \item {\em Theory}: For deep linear networks with orthonormal weight matrices, we prove that local-SSL implements updates identical to those of global BP-SSL, under the conditions in Theorem \ref{th1}.

    \item {\em Theory}: If the orthonormality condition is broken by a shrinking number of neurons for higher layers, we show both theoretically and numerically that adding direct feedback from the last layer makes local-SSL updates more similar to global BP-SSL.
    
    \item {\em Simulation}: In convolutional neural networks, theory indicates that feedback weights should have a structured 2D spatial dependence. Numerically, we show that adding such dependence in local-SSL improves its similarity with global BP-SSL. 

    % \item In convolutional neural networks, we derive that spatially dependent context signal (or, unconstrained feedback weight), and direct feedback from the last layer improve the approximation of BP in theory, and we verify it numerically.
    
    \item {\em Performance on standard data sets}: Our theory driven variations of local-SSL (CLAPP++) improve the classification performance on CIFAR-10, STL-10, Tiny ImageNet and ImageNet (Figure \ref{fig:sketch}D; Table \ref{tab: clapp_compare}). The performance sets a new state-of-the-art performance for local-SSL on these datasets, and reaches that of a comparable global BP-SSL baseline on CIFAR-10, STL-10 and TinyImageNet.

    \item {\em Conceptual}: The generality of our results arises from a novel simplified notation that encompasses multiple recent local-SSL algorithms.% \cite{illing_2021, hinton2022forward, momeni2023phyll, chen2025self}.

\end{itemize}

% \begin{table}[ht]
%     \centering
%     \begin{tabular}{lc}
%     \hline
%     Model   &  STL-10 Accuracy\\
%     \hline
%     \textit{BP-SSL}     & 80.36\\
%     \hline
%     \textit{local-SSL} \cite{illing_2021} & 73.29 \\
%     + unconstrained lateral weights & 78.66\\
%     + direct feedback & 79.38\\
%     \hline
%     \end{tabular}
%     \caption{Performance improvement of local-SSL on STL-10 image classification accuracy after making our theoretically inspired modifications: (1) loosen constraints on lateral weights (red lines in figure \ref{fig:sketch} B and C); (2) introducing the direct feedback in panel C. Green dashed line is end-to-end training of the same self-supervised objective with BP.}
%     \label{tab:summary}
% \end{table}

% Maybe we do not need a separate section level here.
\section{State of the Art}

\paragraph{Local approximation of supervised BP}
Alternatives to supervised BP can be classified into two categories.
The first category approximates or replaces the feedback network, necessary for the backward pass (Figure \ref{fig:sketch}A), by random or trainable feedback weights
 \cite{lillicrap2016fa,akrout2019wm, kunin2020ia}, or by direct feedback from the top layer~\cite{nokland2016dfa,bellec2020eprop,launay2020direct,meulemans2022minimizing,srinivasan2024forward,li2026noprop}. 
The second category uses bidirectionally connected networks. This dynamical system converges to an equilibrium while the output units are successively ``free'' and then ``nudged'' or ``clamped'' to the target value~\cite{lee2015difference,scellier2017equilibrium,Sacramento18,whittington2017predictivecoding,ernoult2019updates,laborieux2021scaling,meulemans2022minimizing, aceituno2023ldifferential,Salvatori2024ipc,max2024contwm}.
A common finding across both categories has been that a better approximation of BP gradients also yields better performance. Below, we apply this philosophy to the case of SSL. 

In principle, some of the BP-approximating methods, like equilibrium propagation or predictive coding, could be adapted for arbitrary loss functions~\cite{scellier2023energy,Salvatori2024ipc}, including those of SSL. Yet, in our hands, naive implementations of these theories did not achieve competitive performance on SSL benchmarks (Appendix \ref{appendix:pc} and Table \ref{tab: clapp_compare}), so we focus on existing local-SSL rules.

\paragraph{Review of local self-supervised learning (local-SSL)}
%REMOVE: Self-supervised learning (SSL) has enabled large scale pre-training of deep networks on large unlabeled datasets. CITATIONS????
Self-supervised learning algorithms can be categorized according to their loss functions as either contrastive (e.g., CPC~\cite{oord2018representation}, SimCLR~\cite{chen2020simclr}, and CLIP~\cite{radford2021learning}); non-contrastive~(e.g., Barlow Twins \cite{zbontar2021barlow} and VICReg \cite{bardes2022vicreg}); or masked input prediction~\cite{bengio2003neural,mikolov2013efficient}.
Here we focus on contrastive learning, which is defined as a classification problem between "positive" and "negative" sample pairs, where positive pairs are data augmentations of the same data sample, and negative pairs come from different data samples.
In non-contrastive methods, negative pairs are not used, but batched neuron activations are regularized to maintain high variance and decorrelation in order to prevent representational collapse.
% Also Define c and z already now? 

The loss of local-SSL is defined per layer and gradients are blocked at every layer. For instance, CLAPP \cite{illing_2021} and LPL \cite{halvagal2023lpl} can be interpreted as layer-wise contrastive (CPC) and non-contrastive (VICReg) SSL, respectively. %They aim to model bioplausible plasticity rules.
Importantly for computational neuroscience, the update steps of both algorithms can be implemented as an extension of Hebbian learning.
%In CLAPP~\cite{illing_2021}, the model hypothesis for biology is that a subject is watching a visual scene and choosing when to fixate or to gaze away toward a new scene. This provides the self-supervision signal between consecutive images that is sufficient to implement a CPC loss variant for binary classification (same object vs. different object, e.g., positive versus negative sample pairs). The weight update could be written as a neo-Hebbian plasticity rule, which depends on inputs in the apical dendrites and a neuromodulator in addition to pre-synaptic and post-synaptic neural activities (Figure \ref{fig:sketch}, B).
%Local, biologically plausible self-supervised learning algorithms could be achieved by optimizing each layer of neurons with layer-specific self-supervised objectives.
%A layer-wise non-contrastive loss similar to LPL \cite{halvagal2023lpl} is also used in 
%\citet{zhu2025stochastic}, though the latter do not aim for biological plausibility. 
So CLAPP and LPL demonstrated that generalized Hebbian learning rules can be stacked to generate hierarchical representations in multi-layer convolutional neural networks. %~\cite{illing_2021, halvagal2023lpl}. 

% \paragraph{Variants of Forward-Forward}
The Forward-forward algorithm \cite{hinton2022forward} is another example of a local contrastive learning algorithm for deep neural networks. A positive sample is a real image from the dataset. A negative sample is a fake image engineered by mixing two different images; or an image with an erroneous label. 
The Forward-forward algorithm maximizes the squared norm of the activations for positive samples and minimizes it for negative samples (it reflects the change of activity in the cortex after surprising events~\cite{homann2022novel}).
The activations are therefore normalized before transmission to the next layer to avoid having a trivial optimization in subsequent layers.
Forward-forward and a variation thereof \cite{papachristodoulou2024convolutional} work on MNIST, but label information is used for scaling to more complex datasets \cite{gong2026adaptive}.
%but handcraft negative samples is not scalable for more complex datasets. 
%\citet{papachristodoulou2024convolutional} have advanced the FF to perform well on CIFAR10, but they rely on label information to design the layerwise objectives.
Self-Contrastive Forward-forward (SCFF)~\cite{chen2025self} enables full self-supervised training 
%using identical or different images as in SimCLR or CLAPP and With additional important engineering improvements, SCFF achieves the 
and sets the present state of the art for local-SSL on image datasets~\cite{chen2025self}. PhyLL \cite{momeni2023phyll} also adapts the local loss of Forward-forward for training physical neural networks.

\begin{table*}[th]
  \caption{\textbf{Comparison of contrastive local-SSL rules.} Choices of $f$, $B^l$, $c_{\rm neg}^l$, and $c_{\rm pos}^l$ for various local-SSL. The reference $\xi^l$ for PhyLL is a fixed vector randomly chosen for each layer, and $\sigma$ is the sigmoid function. In CLAPP, $z'^l$ is the layer activity corresponding to a different augmented copy of the same image that gives $z_{\rm pos}^l$.  The matrix $W^{pred, l}$ is  trained to minimize $\mathcal{L}^l$.}
  \label{tab: binary_contrast}
  \begin{center}
    \begin{small}
      \begin{sc}
        \begin{tabular}{ccccccc}
            \toprule
            learning rule & $c_{\rm neg}^l$ & $c_{\rm pos}^l$ & $B^l$ & Normalization & decreasing function $f(x)$ & loss type \\
            \midrule
            Forward-Forward & $z_{\rm neg}^l$&$z_{\rm pos}^l$ & $I$ & yes & $-x$ & 2\\
            PhyLL & $\xi^l $&$ \xi^l$ & $I$ & yes & $\operatorname{softplus}(-x)$ & 1\\
            SCFF & $z_{\rm neg}^l $&$ z_{\rm pos}^l$ & $I$ & yes &  $-\operatorname{log}\sigma(x - \theta)$ & 2 \\
            CLAPP & $z'^l $&$  z'^l$ & $W^{pred, l}$ & no & $\operatorname{max}(0, 1- x)$ & 2\\
            
            % PhyLL* & $c = z'$ & $W^{pred}$ & no  & $\text{softplus}(x)$ & 1\\
            
            \bottomrule
          \end{tabular}
      \end{sc}
    \end{small}
  \end{center}
  \vskip -0.1in
\end{table*}

\section{Local-SSL approximates global BP-SSL}
\label{sec:theory}

In this section, we first formalize the above contrastive local-SSL methods using a common formalism that will facilitate theoretical analysis. 

%\subsection{Common formalism}
Table \ref{tab: binary_contrast} summarizes how the common formalism can be applied to   CLAPP, Forward-forward, PhyLL, and SCFF.
To define notations, we consider the case of a multi-layer perceptron. Given an $L$-layer network, the activity in each layer $l$ is defined as a real-valued vector $z^{l} = \rho(a^{l})$ where $\rho$ is the non-linear activation function and $a^{l} = W^{l} z^{l-1}$ implements a linear transformation of activities from the previous layer by multiplication with weight matrix $W^{l}$. The input $x$ of the neural network is given as: $z^{0}=x$.

In the context of contrastive algorithms, layerwise activities corresponding to positive and negative input samples are denoted as $z_\text{pos}^l$ and $z_\text{neg}^l$.  They are compared to  reference vectors $c_\text{pos}^l$ and $c_\text{neg}^l$. (For a subset of algorithms,  $c_\text{pos}^l = c_\text{neg}^l$). 
The exact definitions of $c_\text{pos}^l, c_\text{neg}^l$
are different from one algorithm to another (Table \ref{tab: binary_contrast}).

%To train the feedforward weights $ W^{l}$, all the local-SSL algorithms use a self-supervised 
The loss function of all local-SSL algorithms evaluates a scalar score that should be high for $z_\text{pos}^l$ and low for $z_\text{neg}^l$. The scalar score for $z_\text{pos}^l$ is defined as $ {z_\text{pos}^{l} }^\top B^l c_\text{pos}^l$ with a matrix $B^l$; and analogously for $z_\text{neg}^l$. 
Once the two scores have been evaluated, they are combined into a nonlinear loss function $\mathcal{L}^l$ for layer $l$ that takes one of two forms:  
\begin{small}
 \begin{align*}
     \label{def:binary_contrast}
    \mathcal{L}^l = &~ f \left( {z_\text{pos}^{l}}^\top B^l c_\text{pos}^l - {z_\text{neg}^{l}}^\top B^l c_\text{neg}^l \right) & \text{ (type 1) } \\ 
     \mathcal{L}^l = &~ f \left( {z_\text{pos}^{l} }^\top B^l c_\text{pos}^l \right)
     + f \left(- {z_\text{neg}^{l} }^\top B^l c_\text{neg}^l \right) 
      & \text{ (type 2) } % + \lambda || B^l ||^2 ,
 \end{align*}
\end{small}
where $f$ is a decreasing function. The matrix $B^l$ is either trainable in CLAPP-like algorithms or fixed to $I$ in Forward-forward algorithms. Note the different locations of the nonlinearity $f$ when combining the positive and negative score values. For both forms, the essential idea is to ``push'' (in each layer!) the representation of positive and negative inputs into different directions.  Table \ref{tab: binary_contrast} summarizes the definitions of the function $f$
 for different algorithms.

\subsection{Exact equivalence in deep linear networks}
%\subsection{Approximation of end-to-end self-supervised learning in deep linear networks}
For all layerwise local-SSL algorithms, the absence of gradient flow across layers implies that the weight updates in earlier layers are not explicitly optimizing the SSL objective of later layers. Indeed, optimizing all layers with backprop using an SSL loss function in the {\em last} layer (global BP-SSL) leads to better performance than 
local-SSL with the same loss function applied separately in each layer
\cite{illing_2021,halvagal2023lpl} (performance is measured as the accuracy of a downstream linear classifier). 
%both reported that training with layer-wise self-supervised objectives leads to lower decoding accuracy from the last layer when compared to optimizing the same objective on the top layer with end-to-end BP (BP-SSL). 
In this theory section, we examine the mathematical relationship between the gradient update of contrastive local-SSL and the corresponding gradient update of global BP-SSL. 
We will show for deep linear networks that the layerwise optimization with local-SSL and backpropagation gradient update with global BP-SSL  are equal under certain conditions. 
%We also show that the equivalence breaks down as these conditions are relaxed, whereas introducing feedback from the last layer shows less deviation from BP gradients. 

We consider a deep linear network and a loss function of type 2, where $f$ is convex with $f'\le 0$ (the theorems shown below are also valid with losses of type 1, with the same proof technique). %on the positive and negative score $ {z_\text{neg}^l}^\top B^l c^l$ and ${z_\text{neg}^l}^\top B^l c^l$
For the analytical proofs, we added the $L_2$-regularizer using the Frobenius norm $||\cdot||_F$:
\begin{small}
\begin{equation}
\label{eq:loss_lw}
    \mathcal{L}^l = f({z_\text{pos}^l}^\top B^l c_\text{pos}^l) + f(-{z_\text{neg}^l}^\top B^l c_\text{neg}^l) + \lambda ||B^l||_F^2
\end{equation}
\end{small}
The additional regularization enables us to consider a separation of timescales: if $B^l$ is trainable and evolves on a faster timescale than  $W^l$, it is always at the optimal value with respect to the loss.  
Convexity ensures that the corresponding loss $\mathcal{L}_*^l = \operatorname{min}_{B^l} \mathcal{L}^l$ is reached by a unique $B_*^l$.  
Furthermore, we assume orthonormal weight matrices $W^l$, i.e.,  square matrices such that, at the moment of gradient update, the incoming weight vectors to all neurons in any given layer $l$ are orthonormal.
Then we prove the following relationships between local-SSL and global BP-SSL.
Training with global BP-SSL is formalized using the gradient of the loss $\BPL$ at the last layer: $\BPL = \mathcal{L}^L$ and $\BPL_* = \mathcal{L}_*^L$. (To simplify notations, we assume a batch size of one here, and provide the proof for larger batches in Appendix \ref{appendix:proof}):

\begin{theorem}
\label{th1}
We assume a deep linear network with orthonormal weight matrices $W^l$, $l=1,\dots L$. % that are  orthonormal throughout optimization.
Let's consider $\mathcal{L}_*^l$ with $B_*^l=I$ (e.g. forward-forward) or $B_*^l = \operatorname{argmin}_{B^l} \mathcal{L}^l$ (e.g. CLAPP) and $f$ convex.
%For a deep linear network with orthonormal weight matrix $W^l$,
Then gradients of a layer-wise loss $\mathcal{L}_*^l$ are equal to the gradient of $\BPL_*$ backpropagated:
\begin{equation}
    \label{eq:equiv_lw}
    \text{(local-SSL)}~~
    \frac{\partial \mathcal{L}_*^l}{\partial W^{l}_{ij}} = \frac{\partial \BPL_*}{\partial W^{l}_{ij}}
    ~~\text{(global BP-SSL)}
\end{equation}
\end{theorem}
\begin{proof} 
Deriving the backpropagated gradient for the last layer loss we have:
$\frac{\partial \BPL_*}{\partial z^{l}_{pos}} = f'({z_\text{pos}^L}^\top B_*^L c_\text{pos}^L)\cdot (W^L \cdots  W^{l+1})^\top B_*^L c_\text{pos}^L  $.
For the local loss function, we have :
$\frac{\partial \mathcal{L}_*^l}{\partial z^{l}_{pos}} = f'({z_\text{pos}^l}^\top B_*^l c_\text{pos}^l)\cdot B_*^l c_\text{pos}^l $.
The gradients with respect to the parameter are the sum of the outer product of $z^{l-1}$ and these gradients, as well as the opposite contribution of the negative sample.
We can prove theorem \ref{th1} by showing that those two gradients are equal $\frac{\partial \BPL_*}{\partial z^{l}_{pos}}=\frac{\partial \mathcal{L}_*^l}{\partial z^{l}_{pos}}$, which is true if we have the equality $B_*^l = (W^L \cdots  W^{l+1})^\top B_*^L (W^L \cdots W^{l+1})$.
In the case $B_*^l=I$, we simply observe that the products of matrices are cancelling each other because the matrices are orthonormal.
Otherwise, we use the strict convexity of $\mathcal{L}^l$ with respect to $B^l$, which implies that $B_*^l$ is uniquely defined. Then using the orthonormality of $W^l$, it means that the norm of $(W^L \cdots  W^{l+1})^\top B^L (W^L \cdots W^{l+1})$ is the norm of $B^L$. Thus, we have $\BPL(B^L) = \mathcal{L}^l \left((W^L \cdots  W^{l+1})^\top B^L (W^L \cdots W^{l+1})\right)$.
Therefore, if we denote $B_*^L$ the minimum of $\BPL$, this equality implies that the minimum of $\mathcal{L}^l$ is reached for
$B_*^l = (W^L \cdots  W^{l+1})^\top B_*^L (W^L \cdots W^{l+1}) $.
\end{proof}

%\vspace{-0.cm}

{\em Comments}: (i) If there are lateral connections $B^l\ne I$ as in CLAPP, the proof requires $B^l$ to be learnable and its search space to be \textit{unconstrained} and contain $B_*^l$.  %all elements are {\em unconstrained}.
(ii) Orthonormality implies that all layers have an equal number of neurons.
(iii) Linear networks are redundant: if an optimum ($\frac{\partial \BPL_*}{\partial B^L}=0$) is possible in layer $L$, the same optimum is also reachable in layer $l<L$ by optimizing $B^l$. The theorem does not carry over to nonlinear networks. 

\begin{figure}[ht]
    \begin{center}
    
    \centerline{\includegraphics[width=0.9\columnwidth]{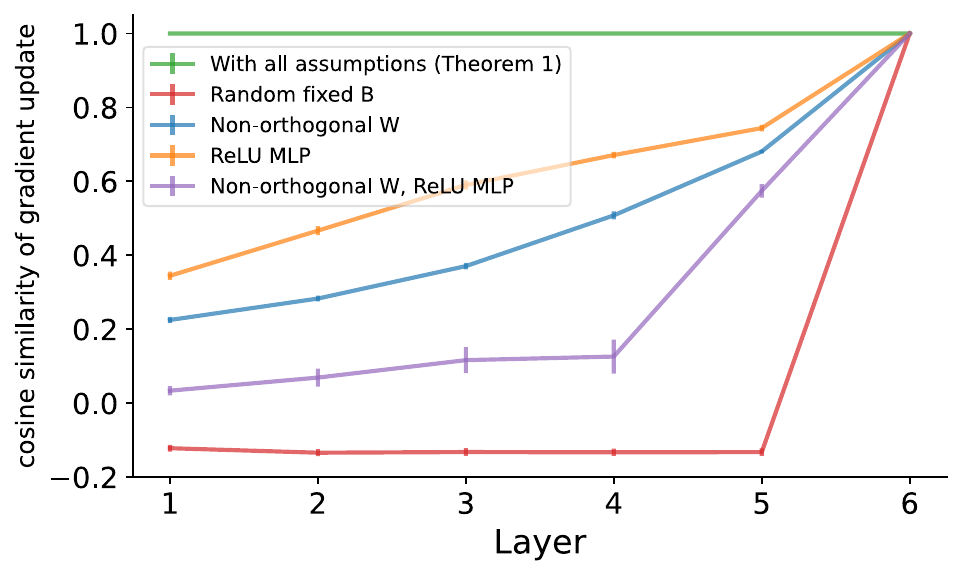}}
    
    \caption{
    \textbf{Numerical verification of Theorem \ref{th1}.}
      Cosine similarity of the gradient update between BP-SSL training and local-SSL rules across layers (error bar indicating 95\% confidence interval computed through different batches of input). Simulations are for theorem \ref{th1} as well as the cases when conditions are dropped. Random fixed B means that the $B^l$ are not optimized to be $B_*^l$. % but rather randomly initialized. 
      ReLU MLP means adding ReLU activation at each layer. Non-orthogonal W means that each $W^l$ is randomly initialized with the default uniform distribution in PyTorch.
    }
    \label{fig_thm1}
    \end{center}
\end{figure}

To verify Theorem \ref{th1} numerically, we implement a 6-layer linear neural network of 128 neurons per layer %(note that orthonormality implies fixed numbers of neurons per layer), 
with feedforward weights $\{ W_l \}_{l=1}^L$ initialized as orthonormal matrices. We compute $B_*^l$ by numerically minimizing the loss $\mathcal{L}^l$  for all layers on a fixed batch of MNIST input samples  (Details in Appendix \ref{appendix:train}). Our simulation confirms that the gradient step of local-SSL and global BP-SSL are numerically identical (Figure \ref{fig_thm1} green line, achieving the maximum cosine similarity of one). 
To study the importance of the different conditions used in the theorem, we remove one condition at a time:
(i) fixed random $B^l$ removes learnability of $B^l$. 
(We also show the effect of adding low-rank constraints to $B^l$ in Figure \ref{fig:rank_constraints});
(ii) non-orthogonal $W$ removes orthonormality of weight matrices;
(iii) ReLU MLP removes linearity. We also study the case of dropping orthonormality and linearity at the same time. 
 As expected, the gradient alignment drops as more conditions are removed, and the drop is more significant for earlier layers (Figure \ref{fig_thm1}).

%This Theorem \ref{th1} defines a form of  \emph{self-consistency} for local-SSL algorithms. 
%Self-consistency breaks, however, if we include the normalization of layer activity as used in FF. The normalization does not affect the computation of the local gradients, but the global BP-SSL gradient will have to include a correction which makes  Theorem \ref{th1} not applicable. Hence, among published local-SSL algorithms, only the ones that are normalization-free (e.g. CLAPP, see Table \ref{tab: binary_contrast}) rigorously enjoy the self-consistency of Theorem \ref{th1}.
Theorem \ref{th1} is applicable to CLAPP in linear networks, but we have not been able to extend it to linear networks that include normalization of layer-activities as required for FF algorithms.
Consequently, we focus our subsequent analysis on CLAPP-like~\cite{illing_2021,Delroq2024critical} algorithms, which are normalization-free and feature a learnable matrix $B^l$. However, our analysis remains valid for any type 1 or type 2 local-SSL loss function, as long as there is a trainable $B^l$ and no normalization.

%\paragraph{Non-linear networks} In practice deep linear network do not compute useful information. So we studied numerically the validity of theory with non-linear MLPs where the theorem does not apply. Figure \ref{fig_thm1} shows that non-linear activation worsen the approximation of the BP gradient.
%Yet we wondered: can we estimate how good can local-SSL can possibly be? And how far are we from this theoretical optimum?
%We could define a theoretical "optimal local-SSL" weight update $\Delta_{\circ}$ that maximally approximates the BP gradients without compromising the locality and the good properties of local-SSL gradient in the case of CLAPP by optimizing $B$ to maximally approximate BP instead of minimizing the local loss:

%\begin{definition}
%We define the \emph{theoretical optimum of local-SSL} update $\Delta_{\circ}$ as the gradient descent update
%$\Delta_{\circ} = \frac{\partial \mathcal{L}^l}{\partial W^{l}} (B_\circ)$ where $B_\circ$ minimizes $\norm{
%    \frac{\partial \mathcal{L}^l}{\partial W^{l}} (B_\circ) - \frac{\partial \mathcal{L}_*^L}{\partial W^{l}}
%   }$.
%\end{definition}

\subsection{Direct feedback in layerwise loss}

%As Figure \ref{fig_thm1} shows that the equivalence in theorem \ref{th1} breaks down as the assumptions are relaxed, we investigate methods to make local-SSL less deviate from BP-SSL gradients. 
%A reasonable intuition is that sending top-down feedback from the last layer would be helpful to approximate the gradient $\frac{\partial \BPL_*}{\partial W^l}$ computed using the activity $z^L$ from the last layer (Figure \ref{fig:sketch}, C).

One strong implication of the orthonormality in Theorem \ref{th1} is that the number of neurons has to be constant across layers. When the network has a number of neurons that reduces across layers, we now show that a modified local-SSL algorithm, where the reference is defined using the last layer $c^l=z'^L$ (Figure \ref{fig:sketch}C), approximates global BP-SSL better than the classic local-SSL algorithm, where $c^l=z'^l$. Here, CLAPP-like algorithms share the same reference for positive and negative samples: $c_\text{neg}^l = c_\text{pos}^l = c^l$. We refer to this algorithm as a local-SSL algorithm with direct feedback (local-SSL DFB) and define it as the gradient update of the loss $\mathcal{L}_{\rm fb}^l$:
%
%\begin{small}
%\begin{equation}
%\label{eq:loss_fb}
%    \mathcal{L}_{fb}^l = f(-{z_\text{neg}^l}^\top D^l c^L)) + f({z_\text{pos}^l}^\top D^l c^L) + \lambda ||D^l||_2^2
%\end{equation}
%\end{small}
%
\begin{equation}
\label{eq:loss_fb}
    \mathcal{L}_{\rm fb}^l = \mathcal{L}^l ~~\text{ with } c^l = z'^L
\end{equation}
Since $c^l$ is considered as constant with respect to $W^l$, the proof of Theorem \ref{th1} is valid under the same assumptions. So when the layer dimension is still constant across layers, we obtain the same result as previously:
\begin{corollary}
    \label{th:equiv_fb}
    For a deep linear network with orthonormal weights $W^l$,
    gradients of a local-SSL with direct feedback are equal to the BP-SSL gradients of $\BPL_*$:
    \begin{equation}
    \text{(local-SSL DFB)}~~
        \frac{\partial \mathcal{L}_{\rm *, fb}^l}{\partial W^l_{ij}}
        =
        \frac{\partial \BPL_*}{\partial W^l_{ij}}
    ~~\text{(global BP-SSL)}
    \end{equation}
\end{corollary}
%
%The proof is the same as for Theorem 3.1.
More interestingly, we now consider a deep network where the number of neurons reduces at each layer.
We assume a deep linear network with semi-orthonormal matrices (meaning that $W^l \in \mathbb{R}^{m^l \times n^l}, m^l < n^l$ and $W^l$ has orthonormal row vectors).
Then Theorem \ref{th1} and Corollary \ref{th:equiv_fb} are not valid anymore. But when $f$ is a linear function, we can prove that local-SSL DFB aligns better with global BP-SSL updates than the standard local-SSL:

\begin{theorem}
    \label{th:low_dim}
    
    Let's consider a deep linear network with semi-orthonormal weights $W^l$ and assume that $f$ is a linear function. 
 Then, assuming batch size one, in comparison to the gradient of global BP-SSL, a local-SSL with direct feedback ($c^l=z^L$) is more similar than the standard local-SSL ($c^l=z^l$):

\begin{small}
\begin{equation}
        \norm{ \frac{\partial \mathcal{L}_{*}^l}{\partial W^l} - \frac{ \partial \BPL_*}{\partial W^l} }_F^2  \geq \norm{\frac{\partial \mathcal{L}_{\rm *, fb}^l}{\partial W^l} - \frac{ \partial \BPL_*}{\partial W^l}}_F^2
\end{equation}
\end{small}
\end{theorem}

% \begin{theorem}
%     \label{th:low_dim}
%     for slides
    
%     Let's consider a deep linear network with semi-orthonormal weights $W^l \in \mathbb{R}^{m^l \times n^l}, m^l < n^l$ and assume that $f$ is a linear function. 
%  Then, assuming batch size one, in comparison to the gradient of global BP-SSL, a local-SSL with direct feedback ($c^l=z^L$) is more similar than the standard local-SSL ($c^l=z^l$):

% \begin{small}
% \begin{equation*}
%         \norm{ \frac{\partial \mathcal{L}_{*}^l}{\partial W^l} - \frac{ \partial \mathcal{L_*^L}}{\partial W^l} }_F^2  \geq \norm{\frac{\partial \mathcal{L}_{\rm *, fb}^l}{\partial W^l} - \frac{ \partial \mathcal{L}_*^L}{\partial W^l}}_F^2
% \end{equation*}
% \end{small}
% \end{theorem}

The proof is written in Appendix \ref{appendix:proof}.

%\begin{proof}
%    (detail in Appendix) The main idea is to decompose the layerwise loss $\mathcal{L}^l$ and $\mathcal{L}_{fb}^l$ using the low-rank subspace spanned by $(W^{L} W^{L-1}\cdots W^{l+1})$, which is introduced by the usage of $z^l$ and $c^l$, and the corresponding null subspace. For a decomposed low-rank subspace, we show that the end-to-end gradient and layerwise gradient are the same. The remaining terms involving the null space make the layerwise gradient deviate from end-to-end training. However, since $\mathcal{L}_{fb}^l$ uses $c^L$ instead of $c^l$, it has fewer components involving null space.
%\end{proof}

\begin{figure*}[ht]
    \begin{center}
    
    \centerline{\includegraphics[width=0.85\linewidth]{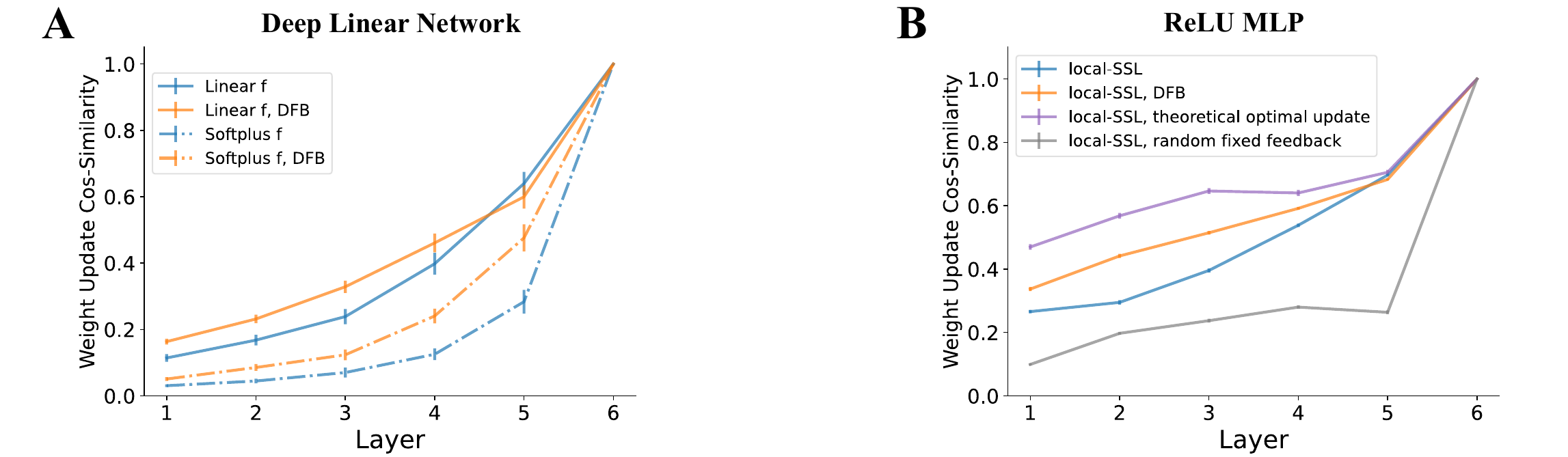}}
    
    \caption{\textbf{Numerical verification that direct feedback improves backpropagation approximation.}
      Cosine similarity of the gradient update between local-SSL and BP-SSL across layers. \textbf{A: } Simulation for theorem \ref{th:low_dim} on a 6-layer deep linear network with shrinking widths, both for linear $f(x)=-x$ and non-linear (softplus) $f(x) = \log(1+\exp(-x))$. \textbf{B: } Comparison of local-SSL and theoretical optimum of BP approximation. Simulations are conducted on a 6-layer ReLU MLP trained with MNIST input. Gray: top-down feedback projection $B^l$ is fixed with random initialization. Purple: train top-down feedback explicitly with definition \ref{def:optimal_fb}. 
    }
    \label{fig_fb}
    \end{center}
\end{figure*}

As previously, we verified the Theorem \ref{th:low_dim} with numerical simulations.
We implement a deep linear network where the dimensionality decreases by half in each layer (from 128 to 4). As shown in Figure \ref{fig_fb}A, when $f$ is linear, the gradient alignment between local-SSL and BP-SSL gradients increases with the introduction of direct feedback. Although the proof only addresses the sum of squared differences for batch size one, the simulation extends to the cosine similarity of the gradient updates for larger batches. We also observe the same trend when $f$ is a softplus function, which is no longer linear but is still a decreasing convex function.

To measure the similarity with the BP-SSL gradients in a more realistic setting, we drop all conditions considered before: we train a 6-layer ReLU MLP with standard initialization for $W^l$ on the MNIST dataset and use batched inputs. 
Additionally, $B^l$ is optimized by gradient descent of local objective $\mathcal{L}^l$ at each iteration, so it does not reach the optimum $B_*^l$ for each input.
Then, we evaluate the cosine similarities with BP-SSL updates on held-out data (Details in Appendix \ref{appendix:train}). Even in this more realistic setting, local-SSL DFB demonstrates a better approximation to BP updates (orange line in Figure \ref{fig_fb}B). As a control, we also train local-SSL with randomly initialized and fixed $B^l$ to confirm that training $B^l$ is important for approximating BP updates (grey line in Figure \ref{fig_fb}B). 

\paragraph{Local-SSL DFB is close to optimal in ReLU MLP}
Since it is impossible to exactly match BP with a single feedback matrix $B^l$ in deep non-linear networks, we also define a theoretical upper bound for BP approximation, which is given by an ``optimal local-SSL" gradient $\Delta_{\circ}$. $\Delta_{\circ}$ is defined as a weight update that has the same form as a local-SSL gradient, but the feedback weights $B^l$ are explicitly optimized to match the BP-SSL gradients by definition \ref{def:optimal_fb}. Although it is possible to make measurements and evaluate $\Delta_{\circ}$ for our analysis, its computation requires BP in practice, so it cannot be local.  We only use this mathematical construction to evaluate an upper bound of the cosine similarity that is achievable between BP-SSL and local-SSL gradients.
\begin{definition}
\label{def:optimal_fb}
We define the \emph{theoretical optimum of local-SSL} update $\Delta_{\circ}$ as the weight update
$\Delta_{\circ} = \frac{\partial \mathcal{L}_{\rm fb}^l}{\partial W^{l}} (B_\circ^l)$ where $B_\circ^l$ is \textbf{not} trained to minimize $\mathcal{L}^l$ but to minimize directly the mean squared error with the gradient of global BP-SSL $\norm{
    \frac{\partial \mathcal{L}_{\rm fb}^l}{\partial z^{l}} (B_\circ^l) - \frac{\partial \BPL_*}{\partial z^{l}}
   }_F^2$ averaged over the dataset.
   %Comment: Not sure but maybe text is lighted than introducing langle and rangle. where $\langle{\cdot}\rangle$ is the average over the dataset.
\end{definition}
To perform a numerical comparison between local-SSL and this theoretical upper bound, we train the ReLU MLP with the ideal $\Delta_{\circ}$ update and then compute the cosine similarity between the local update (local-SSL update with DFB or $\Delta_{\circ}$) and the BP gradients on held-out data (Details in Appendix \ref{appendix:train}).
The narrow margin between local-SSL DFB (orange) and the upper bound (purple) suggests that our local-SSL is closing the gap with this optimal local algorithm. It is non-trivial that local-SSL algorithms would come close to this upper bound because $B^l$ minimizes $
\mathcal{L}^l$ and not directly the similarity with the BP-SSL gradient.

\subsection{Spatially dependent local-SSL for convnets}
\begin{figure*}[ht]
    \begin{center}
    
    \centerline{\includegraphics[width=0.95\linewidth]{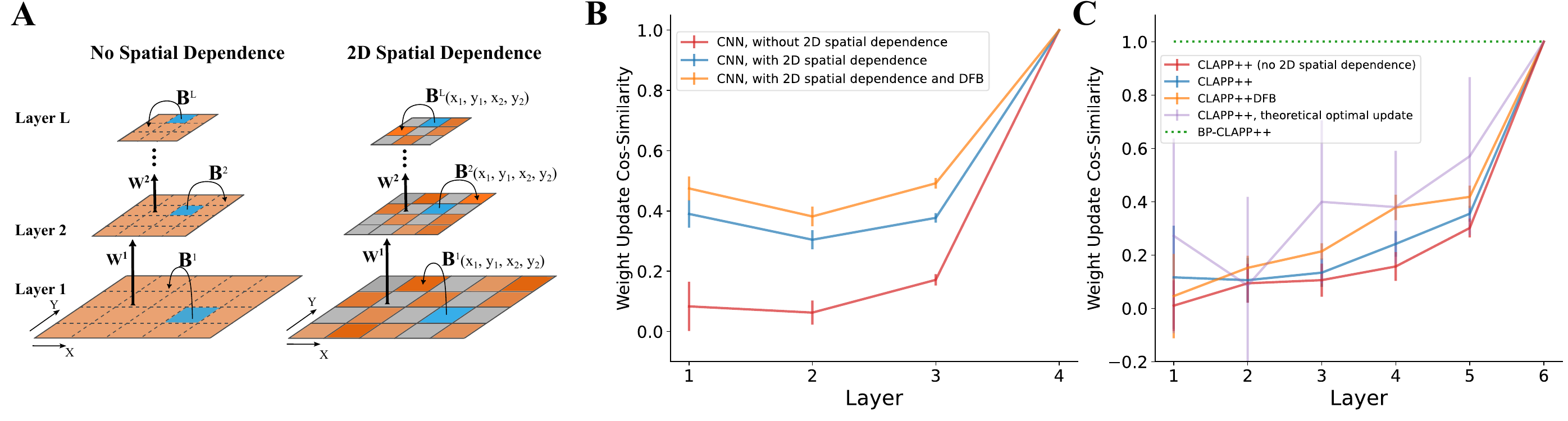}}
    
    \caption{\textbf{Numerical verification that spatial dependence improves BP approximation.}
      \textbf{(A):} Illustration of spatial dependence: without spatial dependence (left), the same $B^l$ is used to project $c^l$ (blue) onto neurons across the feature map, so they share the same color; with spatial dependence (right), lateral projections $B^l$ depend on the locations $(x_1, y_1) \text{ and } (x_2, y_2)$, of $z^l$ and $c^l$ in their layer's feature map. Only nearby neurons with the same color share the same projection $B^l$. %As the spatial dimension decreases in later layers, the width of patches decreases (Exact numbers in Appendix \ref{appendix:train}).
      \textbf{(B) and (C): }   Cosine similarity of the gradient update between BP-SSL and local-SSL across layers. 
      %Red: $B^l$ projects $c^l$ from the same layer, no spatial dependence. Blue: $B^l$ projects $c^l$ from the same layer, with 2D spatial dependence. Orange: $B^l$ projects $c^L$ from the last layer as feedback, with 2D spatial dependence.
      For \textbf{B}, values are obtained from 4-layer linear convnets with optimal $B_*^l$. Kernel and stride have length 2 for simplicity. For \textbf{C}, values are obtained from VGG models trained on STL10. Purple: theoretical optimal update $\Delta_\circ$ obtained from training following the definition \ref{def:optimal_fb}. (Details in Appendix \ref{appendix:train}). The plotted measurements are performed after 100 epochs of training. 
    }
    \label{fig_cnn}
    \end{center}
\end{figure*}

\begin{table*}[ht]
  \caption{\textbf{Local-SSL achieves the performance of BP-SSL.} Different local-SSL rules tested on various image classification benchmarks. The BP-SSL references are named BP-CLAPP++ when the same loss and network of CLAPP++ is used, but the gradients are propagated end-to-end by BP. In BP-InfoNCE, we replaced the CLAPP++ loss with the InfoNCE loss~\cite{oord2018representation, chen2020simclr}. Naive CLAPP DFA/Predictive Coding are the performance of using DFA \cite{nokland2016dfa} and Predictive Coding \cite{Salvatori2024ipc} to train the last layer objective $\BPL$ of CLAPP. When the citation is not provided, the results are produced from our own simulations on the same VGG convnet setup. Values after $\pm$ are the 95\% confidence interval computed from 4 simulations of different random seeds. }
  \label{tab: clapp_compare}
  \begin{center}
    \begin{small}
          \begin{tabular}{lcccccc}
            \toprule
             &  Local & 2D Spatial & CIFAR10 & STL-10 & Tiny-ImageNet & ImageNet \\
            Method & update & Dependence & accuracy & accuracy & accuracy & accuracy\\
            \midrule
            BP-CLAPP++ &  no & - & 80.49 $\pm$ 0.20 & 80.36 $\pm$ 0.26 & {37.55 $\pm$ 0.61} & 48.52\\
            BP-InfoNCE & no & - & 80.69 $\pm$ 0.84 & 81.97 $\pm$ 1.23 & 36.78 $\pm$ 0.73 &   55.19 \\
            \midrule
            Naive CLAPP DFA &  yes & - &  - & 52.30 & - & - \\
            Naive CLAPP Predictive Coding &  yes & - & 46.52 & 36.75 & - & - \\
            \midrule
            CLAPP \cite{illing_2021} &  yes & no & - & 73.6 & - & - \\
            LPL \cite{halvagal2023lpl} &  yes & no & 59.4 & 63.2 & - & - \\
            SoftHebb \cite{journe2022hebbian} & yes & no & 80.31 & 76.23 & - & 27.3 \\
            Stochastic FF \cite{zhu2025stochastic} &  yes & no & 76.96 & - & - & - \\
            SCFF \cite{chen2025self} & yes & yes & 80.60 & 77.14 & 35.67& - \\
            CLAPP++ (no 2D spatial dependence) &  yes & no & 73.21 & 75.10 & 28.18 & 38.31\\
            CLAPP++ & yes & yes & 80.51 $\pm$ 0.27 &78.66 $\pm$ 0.16 & 36.63 $\pm$ 0.32 & 42.55\\
            CLAPP++DFB & yes & yes & 80.65 $\pm$ 0.18	& 79.38 $\pm$ 0.17	& 36.70 $\pm$ 0.24 & \textbf{44.16}\\
            CLAPP++both & yes & yes &\textbf{81.18 $\pm$ 0.18}	& \textbf{79.62 $\pm$ 0.28}	& \textbf{37.78 $\pm$ 0.67} & 42.49 \\
            \bottomrule
        \end{tabular}
    \end{small}
  \end{center}
\end{table*}

Beyond ReLU MLP, we intend to design a local-SSL algorithm that is competitive with CLAPP and SCFF with convnet architectures.
We follow the same approach and seek algorithmic modifications that would improve the gradient approximation of BP-SSL in the convnet architecture.
We explain below how we found that adding spatial structure to the loss $\mathcal{L}^l$ was important in our case.

In convnets, $z^l$ are 2D feature maps and the feedforward matrix product $W^l z^{l-1}$ is replaced with a 2D convolution $W^l \ast z^{l-1}$. % start of rebuttal below
In previous local-SSL algorithms such as CLAPP, LPL, or GIM (block-wise SSL) \cite{lowe2019gim}, the local losses are computed after averaging the feature $z^l, c^l$ over all spatial locations, which can be defined as follows with our notations (the global average pooling operator denoted $\gpool$ averages the feature maps over the spatial dimensions):
\begin{align}
    {\cal L}^l = \ &f(\gpool(z^l)^T B^l \gpool(c^l))
    \label{eq:global_pool}
\end{align}
For brevity, we only write the contribution of a single sample pair $(z^l, c^l)$, but the pooling operator applies similarly to both positive and negative sample pairs.  
With global average pooling, the gradient $\partial {\cal L}^l /\partial z^l$ is shared across the spatial dimensions. This is in contradiction with the ideal projection weight $\partial \BPL /\partial z^l$ which is not shared across the feature maps with BP. Instead, we define an alternative such that $B^l$ may learn cross dependence between spatial locations of $c^l$ and $z^l$. Local-SSL with 2D spatial dependence can be implemented as follows ($\pool_k$ means averaging within patches of size $k$): 
\begin{align}
    {\cal L}^l = \ & f\left(\flatten(\pool_{k_1}(z^l))^T B^l \flatten(\pool_{k_2}(c^l))\right) 
    \label{eq:spatial_dependence}
\end{align}
A similar implementation has also been studied for blockwise training in \citep{siddiqui2024blockwise}.
In comparison with equation \eqref{eq:global_pool}, with 2D spatial dependence the matrix $B^l$ now has more parameters in equation \eqref{eq:spatial_dependence}. Now the size of $B^l$ increases linearly with the size of the image, in such a way that the gradient $\partial {\cal L}^l / \partial z^l$ is only shared within a patch of size $k_1$ instead of the entire image in equation \eqref{eq:global_pool} (Figure \ref{fig_cnn}A).
Ideally the patch sizes $k_1$ and $k_2$ would be as small as possible ($k_1=k_2=1$), but this may make the VRAM requirement prohibitive. In particular, this method is able to compute exact BP gradients in the ideal scenario:
\begin{proposition}
\label{prop:spatial_dependence}
    We assume a deep linear convolutional neural network (without non-linearity and max-pooling).
    For local-SSL with 2D spatial dependence (equation \ref{eq:spatial_dependence}) and patch size $k_1=k_2=1$, there exists $B^l$ such that local-SSL and BP-SSL gradients are exactly equivalent: $\partial {\cal L}^l/\partial W_{ij}^l = \partial {\bf L}/\partial W_{ij}^l$. 
\end{proposition}
The proof uses the fact that the search space of $B^l$ includes all possible linear operators between the flattened feature maps $c^l$ and $z^l$, which contains the composition of convolutions required to compute the BP gradients. Details are provided in Appendix \ref{appendix:proof}.
In contrast, without spatial dependence (equation \ref{eq:global_pool}), there exists cases where sharing $B^l$ across all spatial locations is unable to compute the BP gradients.

% Original Text
%To compute the value ${z^l}^\top B^l c^l$ in local objectives $\mathcal{L}^l$, \citet{illing_2021} first average the feature $z^l, c^l$ over all spatial locations. However, it means that $B^l$ is independent of the 2D spatial locations of the feature map $z^l$ and $c^l$, leading to a constrained search space of $B^l$.

%We construct instead a linear operator $B^l$ that reflects the structured 2D spatial dependence of the BP-SSL gradient. BP errors from the last layer are propagated through the successive backward operators of the convolution $W^l \ast z^{l-1}$ and non-linearities.
%This results in a BP-SSL gradient that is spatially dependent, with different gradients for different spatial locations of the feature maps. Following this intuition, we use different $B^l$ depending on the spatial locations of $z^l$ and $c^l$ in each feature map (Figure \ref{fig_cnn}A), and nearby features share the same $B^l$ to reduce the memory cost.
To verify in simulations that the 2D spatial dependence improves the alignment with the ideal BP gradients, we consider a 4-layer linear convnet (details in Appendix \ref{appendix:train}) and find that 2D spatial dependences improve the alignment with BP-SSL gradients (Figure \ref{fig_cnn}B). Consistently with our previous results we also see empirically that using spatially dependent feedback in combination with DFB from the last layer further improves the gradient alignment.

\section{Empirical results} 
\label{sec:empirical_results}

Although our analysis shows that local-SSL is almost the ideal approximation of BP-SSL update, the gap in cosine similarity in early layers may appear substantial.
The decisive test is to measure whether local-SSL matches the functional performance of global BP-SSL pre-training. We train the same convolutional networks as \citet{illing_2021}. The network is first pre-trained using different learning rules. To measure the quality of the learned representation, we freeze the network and report the accuracy of a supervised linear classifier (details in Appendix \ref{appendix:train}). 

%\subsection{Layerwise contrastive learning performs close to end-to-end training}

% Inspired both by the recent advancement in self-supervised learning and local learning rules and by our analysis on gradient alignments, we made the following modifications to the original CLAPP \cite{illing_2021} model:
% \begin{enumerate}

%     \item Following the practice of variants of the Forward-Forward algorithms \cite{hinton2022forward, chen2025self}, we concatenate representations from multiple layers to train the decoder.
    
%     \item Following the setup of SimCLR \cite{chen2020simclr} we use data augmentations to create positive and negative contrastive samples. In contrast, the original CLAPP \cite{illing_2021} uses non-overlapping patches from the same image as positive samples and patches from different images as negative samples.
    
%     \item We introduce the spatial dependence by not averaging the spatial dimension as done in \citet{illing_2021}. To find a balance between retaining spatial information and reducing the network size, we still perform local spatial averaging (Details in Appendix B).
% \end{enumerate}

Before studying the changes of the algorithm guided by our theory, we build a baseline upon the original CLAPP \cite{illing_2021}.
We made two notable modifications from the original CLAPP. First, we used the data augmentations in SimCLR \cite{chen2020simclr} rather than non-overlapping image patches to create contrastive samples. Second, after self-supervised training, we read-out concatenated representations from multiple layers to train the linear classifier.

\begin{figure}[ht]
    \begin{center}
    
    \centerline{\includegraphics[width=0.9\columnwidth]{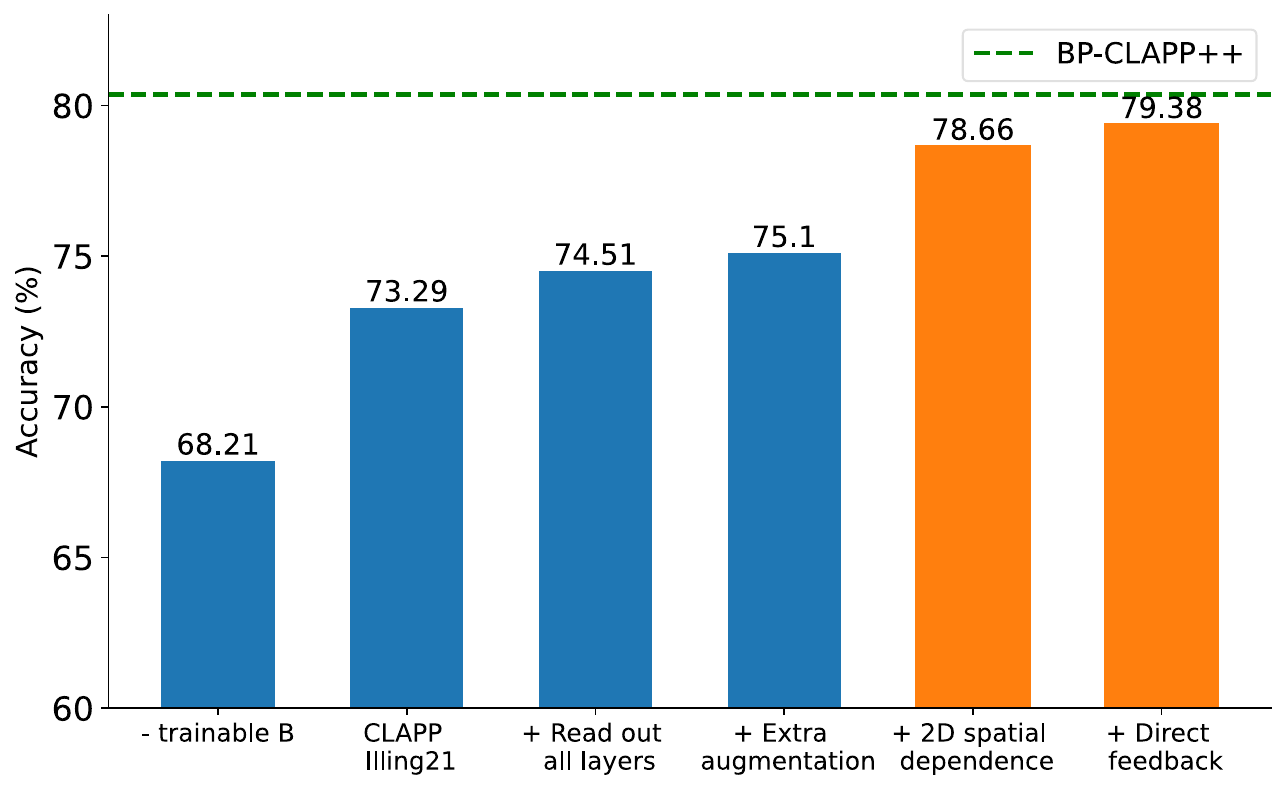}}
    
    \caption{
      Comparison between CLAPP \cite{illing_2021} and CLAPP++ performance on STL10. Orange bars are theory guided algorithmic changes. %, Blue corresponds to algorithmic changes borrowed from the literature.% The "Extra augmentations" are taken from the SimCLR paper \cite{chen2020simclr}. 
    }
    \label{fig_ablation}
    \end{center}
\end{figure}

\paragraph{Spatial dependence}
Then, we add spatial dependence to the projection $B^l$, based on the analysis in the previous section. We call this new learning rule CLAPP++, as it uses the same objective function as CLAPP. Consistent with our theory and simulation in linear networks, we find that the usage of spatial dependence makes the gradient of the local-SSL more similar to the BP-SSL learning algorithm during training of non-linear convnets (Figure \ref{fig_cnn}C). Ablation studies of CLAPP++ demonstrate that the most significant performance improvements on CIFAR10, STL10, TinyImageNet and ImageNet originate from the introduction of spatial dependence (Table \ref{tab: clapp_compare}, Figure \ref{fig_ablation}).

\paragraph{Direct feedback}
Next, we introduce direct top-down feedback and train a model CLAPP++DFB, which is implemented by defining reference input from the final layer $c^l=z^L$. Similar to our results on ReLU MLP, the alignment between CLAPP++DFB gradient and BP gradient further increases, and the gradient of CLAPP++DFB is also close to the `optimal' gradient approximation obtained by explicitly training the feedback with the update in definition \ref{def:optimal_fb} (Figure \ref{fig_cnn}, purple line). The improvements in gradient alignment are also consistent throughout different stages of training (Figure \ref{fig:grad_by_ep}). Across the four datasets, CLAPP++DFB appears to provide a small but consistent improvement of performance (Figure \ref{fig_ablation}, Table \ref{tab: clapp_compare}). We also train a model CLAPP++both, in which we optimize the sum of two losses $\mathcal{L}_{\rm fb}^l + \mathcal{L}^l$ at each layer. This model achieves the best performance on CIFAR10, STL10, and Tiny ImageNet dataset (Table \ref{tab: clapp_compare}).

\paragraph{Local-SSL SOTA}
As summarized in Table \ref{tab: clapp_compare}, we compared our results to the layer-wise SSL (or unsupervised) algorithms that we could find in the literature. Previously, the best performing algorithm was Self-Contrastive Forward Forward (SCFF) or SoftHebb \cite{journe2022hebbian} depending on the dataset. All our CLAPP++ variants with theoretically guided spatial dependence reach a higher accuracy on all datasets. It therefore appears that CLAPP++ is the current state-of-the-art of local-SSL algorithms.

\begin{table*}
  \caption{\textbf{Estimation of VRAM and training time requirements with ideal Local-SSL algorithms}: peak RAM is the total memory allocated to store layer activations. Wall clock time is the averaged time used per batched iteration on a A100 GPU. All values are computed based on Algorithms \ref{alg:ideal} or \ref{alg:ideal_dfb} using a batch size of 256.}
  \label{tab: efficiency}
  \begin{center}
    \begin{small}
          \begin{tabular}{lcccc}
            \toprule
               & \multicolumn{2}{c}{STL-10 with 6 layer VGG}  &  \multicolumn{2}{c}{ImageNet with 8 layer VGG}  \\
              &  &  &  & \\
            Method  & Peak VRAM (GB) & Wall Clock (ms) & Peak VRAM (GB) & Wall Clock (ms)\\
            \midrule
            BP-CLAPP++ &   5.47 & 146 & 11.10 & 384\\
            CLAPP++ (no 2D spatial dependence) & \textbf{3.38} & \textbf{125} & \textbf{4.60} & \textbf{330} \\
            CLAPP++ &  \textbf{3.38} & 146 & \textbf{4.60} & 376\\
            CLAPP++DFB &  3.45	& 141 & 4.69  & 360\\
            \bottomrule
        \end{tabular}
    \end{small}
  \end{center}
\end{table*}

\paragraph{Local-SSL reaches the performance of BP-SSL.}
Reaching the performance of backpropagation has been notoriously hard with local supervised learning rules~\cite{akrout2019wm,launay2020direct}, and it was also never achieved in SSL setups.
For our CLAPP++ variants with spatial dependence (CLAPP++, CLAPP++DFB, CLAPP++both), we find that local-SSL brings downstream classification accuracies close to the performance of BP-SSL on CIFAR-10, STL-10 and TinyImageNet. The most comparable BP-SSL baseline is BP-CLAPP++, where the CLAPP++ loss function of the last layer is optimized end-to-end with BP (BP-CLAPP++ in Figure \ref{fig_ablation}, Table \ref{tab: clapp_compare}). The margin in classification accuracy has disappeared between our local CLAPP++ and BP-CLAPP++ on all datasets with the exception of ImageNet which is harder and has a higher image resolution. To further show that this achievement is not trivially caused by the limitation of the SSL loss used by CLAPP++, we also train the InfoNCE loss, which is used in CPC, SimCLR and CLIP \cite{oord2018representation,chen2020simclr,radford2021learning}. As shown in Table \ref{tab: clapp_compare}, our local CLAPP++ algorithm also matches the performance of BP-InfoNCE on CIFAR-10, STL-10 and Tiny ImageNet. 
Note that our BP-InfoNCE baseline is not, however, the state-of-the-art of global BP-SSL. It uses linear projection heads and the same VGG architecture as CLAPP++. Better performance is expected with non-linear projection heads and ResNet architecture~\cite{chen2020simclr}.

\paragraph{Reducing VRAM requirements with local-SSL.}
Given a large batch size, the GPU VRAM is mostly saturated during BP training by the storage of the network activations.
With local-SSL, it is not necessary to store the whole network activations as we may apply the local gradients immediately after the forward pass of each individual layer. Hence, unlike with BP, the peak VRAM allocation does not scale with network depth. Instead it is dominated by the size of the largest layer. With our architectures and datasets, we estimate that CLAPP++ saves between 38\% to 60\% peak VRAM for storing layer activations with batch size 256 in comparison with BP-SSL (Table \ref{tab: efficiency}).
We estimated the peak VRAM required for local-SSL as two tensors of float32 activations, the largest layer $z^l$ and its input $z^{l-1}$. For BP-SSL we estimated that we store one float32 activation tensor per layer.
At this point we did not perform an entire training with the ideal algorithms that take advantage of this feature of local-SSL (see Algorithm \ref{alg:ideal} and \ref{alg:ideal_dfb} in Appendix). Instead for the benchmark datasets we used a mathematically equivalent alternative implemented with stop gradient (Algorithm \ref{alg:detach} in Appendix).

\section{Discussion}

The central theoretical contribution is to show that some local-SSL algorithms can implement exactly BP-SSL in deep linear networks.
This theoretical finding has guided our improvements of local-SSL algorithms to increase the similarity with BP-SSL and the algorithm performance.
%In particular, direct top-down feedback and 2D spatial dependence can increase the similarity with BP-SSL. 
%Our local-SSL algorithm CLAPP++ sets a new state-of-the-art performance for local-SSL algorithms across three image datasets, and approaches the performance of BP.

% So far, we have not exploited the resource efficiency of local-SSL.
%Besides the potential with custom hardware~\cite{su2025elfcore,momeni2023phyll}, CLAPP++DFB has a straightforward advantage over BP and CLAPP: instead of storing the activation or the context vector $c^l$ at every layer, CLAPP++DFB requires a single context vector $z^L$ for the whole network, which reduces the memory footprint. We leave a deeper analysis to future work. 

While the scale of our experiments are consistent with the standards of local learning literatures, we acknowledge that mainstream self-supervised learning models have scaled to more complex architectures such as ResNets and larger datasets. Bridging this gap will require addressing non-trivial challenges such as adapting local learning rules to residual connections, which would be an important direction for future works.

\paragraph{Biological plausibility of CLAPP++}
Local learning rules like CLAPP \cite{illing_2021} and LPL \cite{halvagal2023lpl} were intended to model plasticity in the brain.
For CLAPP, the bio-plausibility argument comes from the Hebbian-like weight update. For the matrix $B^l$ the update is a classical hebbian rule $\Delta B_{jk}^{l} = \gamma \cdot z_j^{l} \cdot c_k^{ l}$ where $\gamma\in \{-1,0,1\}$ is reminiscent of the effect of neuromodulators. The update of the feedforward weight $W^l$ has a more important role with CLAPP: the plasticity is additionally gated by the contextual activity $c^l$ and can be interpreted as the apical dendritic input that is thought to gate plasticity \cite{larkum1999burst,bittner2017behavioral,illing_2021}:
\begin{equation}
    \label{eq:clapp_fb_rule}
    \Delta W_{ji}^{l} =
    \underbrace{\gamma}_{\text{neuromodulator}}
    \cdot
    \underbrace{(B^{l} c^{l})_j}_{\text{dendritic prediction}}
    \cdot
    \underbrace{\rho'(a_j^{l}) z_i^{l-1}}_{\text{hebbian term}}~.
\end{equation}
%\begin{equation}
%    \label{eq:clapp_fb_pred}
%    
%\end{equation}
In CLAPP++, this interpretation is not compromised, and it even elucidates a question that remained in CLAPP. Apical dendrites in the cortex are thought to integrate inputs from distant higher brain areas \cite{harris2015neocortical,gilbert2013top, marques2018functional, keller2020fbvisual,nejad2025sslfb}.
Recent \textit{in vitro} studies confirmed that apical dendrites modulate the synaptic plasticity in basal dendrites \cite{aceituno2024apical}.
As illustrated in Figure \ref{fig:sketch}C, this is better captured with CLAPP++DFB, where the learning signal comes from the top layer while the dendritic prediction $B^l c^l$ comes from nearby neurons in CLAPP.
Similarly, our 2D spatial dependence model may explain some of the structured connectivity visible across the visual cortex \cite{harris2019hierarchical}.

More generally, our study shows that the gradient of global SSL principles can be well approximated by local biologically plausible learning rules. Thereby, it lifts the requirement for the contested machinery of backpropagation to implement efficient gradient based representation learning.

%Also, in terms of temporal locality, the interpretation from CLAPP \cite{illing_2021} is that $c^L$ represents neural activity of the input received at an earlier time. This delay in time would enable the feedforward $z^{l}$ and the feedback $B^lc^L$ to arrive at layer $l$ together.

%\paragraph{Comparisons with other methods with direct feedback} Our work is fundamentally built upon the layerwise optimization of contrastive objectives as proposed in CLAPP \cite{illing_2021}, while also introducing spatially dependent direct feedback signals from the final output layer for synaptic update. The direct feedback structure is similar to Direct Feedback Alignment \cite{nokland2016dfa}, but the feedback weights are trained by a Hebbian learning rule. \citet{meulemans2022minimizing} also trains a direct feedback with Hebbian learning rules, but the plasticity relies on the filtered neural response to a high frequency noise rather than the neural activity itself, which adds an extra complexity on bio-plausibility and computational efficiency. \citet{srinivasan2024forward} uses direct feedback from the last layer to create negative inputs for the forward-forward algorithm. In contrast to their method, our feedback structure projects network activity from the output layer to every intermediate layer, regardless of whether the input is positive or negative (Detailed comparison in Appendix \ref{appendix:clarify}).

\newpage

\section*{Impact Statement}

This paper presents work whose goal is to advance the field of neuroscience and machine learning research. There are no potential societal consequences of our work
that we feel must be specifically highlighted here.

\section*{Acknowledgments}
This research is supported by the Swiss National Science Foundation grants (200020-207426 and 200021-236436), Vienna Science and Technology Fund (VRG24-018) and Machinita Technologies. We thank Tommaso Salvatori, Zhe Su, Brian Yuan, and everyone from the laboratory of computational neuroscience (LCN) at EPFL for their support and discussion. We also especially want to thank Xing Chen and Bernd Illing for their GitHub code on previous work and for providing deep and insightful feedback on the paper.

% In the unusual situation where you want a paper to appear in the
% references without citing it in the main text, use \nocite

\bibliography{References}
\bibliographystyle{icml2026}

%%%%%%%%%%%%%%%%%%%%%%%%%%%%%%%%%%%%%%%%%%%%%%%%%%%%%%%%%%%%%%%%%%%%%%%%%%%%%%%
%%%%%%%%%%%%%%%%%%%%%%%%%%%%%%%%%%%%%%%%%%%%%%%%%%%%%%%%%%%%%%%%%%%%%%%%%%%%%%%
% APPENDIX
%%%%%%%%%%%%%%%%%%%%%%%%%%%%%%%%%%%%%%%%%%%%%%%%%%%%%%%%%%%%%%%%%%%%%%%%%%%%%%%
%%%%%%%%%%%%%%%%%%%%%%%%%%%%%%%%%%%%%%%%%%%%%%%%%%%%%%%%%%%%%%%%%%%%%%%%%%%%%%%
\newpage
\appendix
\onecolumn
\setcounter{figure}{0}
\setcounter{table}{0}
\renewcommand{\thefigure}{A\arabic{figure}}
\renewcommand{\thetable}{A\arabic{table}}

\section{Local-SSL Algorithms}

Algorithm \ref{alg:ideal} is the ideal way of implementing local-SSL without direct feedback, and Algorithm \ref{alg:ideal_dfb} is the ideal way of implementing local-SSL with direct feedback.
In our implementation, we adopt Algorithm \ref{alg:detach}, in which we use the stop gradient function to disrupt the gradient flow and then use the autogradient function in PyTorch.

\begin{algorithm}[h]
  \caption{Ideal algorithm for local-SSL}
  \label{alg:ideal}
  \begin{algorithmic}
    %\STATE {\bfseries Input:} data $x_{pos},\ x_{neg},\ x'$
    %\STATE {\bf Initialize:} $z_\text{pos}^0 = x_{pos},\ z_\text{neg}^0 = x_{neg},\ c^0 = x'$
    \FOR{$l=1$ {\bfseries to} $L$}
    \STATE $z_\text{pos}^l = \text{Layer}_l(z_\text{pos}^{l-1}),\ z_\text{neg}^l = \text{Layer}_l(z_\text{neg}^{l-1}),\ c^l = \text{Layer}_l(c^{l-1})$
    \STATE {Compute} $\nabla_{W^l}\mathcal{L}^l$ using $z_\text{pos}^l, z_\text{neg}^l, B^l, c^l$
    \STATE {Update} $W^l$ with $\nabla_{W^l}\mathcal{L}^l$
    \STATE {\bf Delete} $z_\text{pos}^{l-1},\ z_\text{neg}^{l-1},\ c^{l-1}$
    \ENDFOR
  \end{algorithmic}
\end{algorithm}

\begin{algorithm}[h]
  \caption{Ideal algorithm for local-SSL with DFB}
  \label{alg:ideal_dfb}
  \begin{algorithmic}
    %\STATE {\bfseries Input:} data $x_{pos},\ x_{neg},\ x'$
    %\STATE {\bf Initialize:} $z_\text{pos}^0 = x_{pos},\ z_\text{neg}^0 = x_{neg},\ c = x'$
    \FOR{$l=1$ {\bfseries to} $L$} 
    \STATE $c = \text{Layer}_l(c)$
    \ENDFOR
    \FOR{$l=1$ {\bfseries to} $L$}
    \STATE $z_\text{pos}^l = \text{Layer}_l(z_\text{pos}^{l-1}),\ z_\text{neg}^l = \text{Layer}_l(z_\text{neg}^{l-1})$
    \STATE {Compute} $\nabla_{W^l}\mathcal{L}^l$ using $z_\text{pos}^l, z_\text{neg}^l, B^l, c$
    \STATE {Update} $W^l$ with $\nabla_{W^l}\mathcal{L}^l$
    \STATE {\bf Delete} $z_\text{pos}^{l-1},\ z_\text{neg}^{l-1}$
    \ENDFOR
  \end{algorithmic}
\end{algorithm}

\begin{algorithm}[h]
  \caption{Implemented algorithm for local-SSL using stop gradients (SG)}
  \label{alg:detach}
  \begin{algorithmic}
    %\STATE {\bfseries Input:} data $x_{pos},\ x_{neg},\ x'$
    \FOR{$l=1$ {\bfseries to} $L$}
    \STATE  $z_\text{pos}^l = \text{Layer}_l(SG(z_\text{pos}^{l-1})),\ z_\text{neg}^l = \text{Layer}_l(SG(z_\text{neg}^{l-1})), c^l = \text{Layer}_l(SG(c^{l-1}))$
    \ENDFOR
    \STATE {\bf Initialize:} $\mathcal{L}_{tot} = 0$
    \FOR{$l=1$ {\bfseries to} $L$}
    \STATE {Compute} ${\cal L}^l$ using $z_\text{pos}^l, z_\text{neg}^l, B^l, c^l$ (or $c^L$ with DFB)
    \STATE $\mathcal{L}_{tot} = \mathcal{L}_{tot} + \mathcal{L}^l$
    \ENDFOR
    \STATE {Update} all $W^l$ with $\nabla_{W^l}\mathcal{L}_{tot} $
    
  \end{algorithmic}
\end{algorithm}

\section{Proof}
\label{appendix:proof}

\subsection{Proof of batched theorem \ref{th1} with batched loss function}

Consider the loss of type 2 with batched input $\mathcal{B}$ and L-2 regularization:
 \begin{align*}
     \mathcal{L}^l = \sum_{\mu\in \mathcal{B}} f \left(- {z_\text{neg}^{\mu, l} }^\top B^l c^{\mu, l} \right) + f \left( {z_\text{pos}^{\mu, l} }^\top B^l c^{\mu, l} \right) + \lambda || B^l ||_F^2 ,
 \end{align*}
 We show as in the Theorem 3.1 that $\frac{\partial \mathcal{L}^l}{\partial W_{ij}^l} = \frac{\partial \mathcal{L}^L}{\partial W_{ij}^l}$ but now each loss has a batch dimension.
The core of the proof is identical to the batch-size-one version described in the main text because there is no interdependencies across batch dimensions. 

\begin{proof}
Deriving the backpropagated gradient for the last layer loss we have:
$\frac{\partial \BPL_*}{\partial z^{\mu, l}_{pos}} =f'((z_\text{pos}^{\mu, L})^\top B_*^L c^{\mu, L})\cdot (W^L \cdots  W^{l+1})^\top B_*^L c^{\mu, L}  $.
For the local loss function we have :
$\frac{\partial \mathcal{L}_*^l}{\partial z^{\mu, l}_{pos}} = f'((z_\text{pos}^{\mu, l})^\top B_*^l c^{\mu, l})\cdot B_*^l c^{\mu, l} $.
The gradients with respect to the parameter are the sum of the outer product of $z^{l-1}$ and these gradients scaled by local derivatives, and the opposite contribution of the negative sample. Because $\frac{\partial \BPL_*}{\partial W_{ij}^l} = \sum_\mu (\frac{\partial \mathcal{L}_*^L}{\partial z^{\mu, l}_{pos, i}} \frac{\partial z^{\mu, l}_{pos, i}}{\partial z^{\mu, l-1}_{pos, j}} + \frac{\partial \mathcal{L}_*^L}{\partial z^{\mu, l}_{neg, i}} \frac{\partial z^{\mu, l}_{neg, i}}{\partial z^{\mu, l-1}_{neg, j}})$, the difference between local-SSL gradients and BP gradients would only appears in the partial derivative of loss $ \BPL_*$ with respect to layer activations. Therefore, we can prove batched theorem \ref{th1} by showing that those two gradients are equal $\frac{\partial \BPL_*}{\partial z^{\mu, l}_{pos}}=\frac{\partial \mathcal{L}_*^l}{\partial z^{\mu, l}_{pos}}$, which is true if we have the equality $B_*^l = (W^L \cdots  W^{l+1})^\top B_*^L (W^L \cdots W^{l+1}) $. (This claim could be shown by simply substituting the equality into $\frac{\partial \mathcal{L}_*^l}{\partial z^{\mu, l}_{pos}} = f'((z_\text{pos}^{\mu, l})^\top B_*^l c^{\mu, l})\cdot B_*^l c^{\mu, l} $.) We could then repeat the same method as the one in the proof for batch 1.
In the case $B_*^l=I$, we simply observe that products of matrices are cancelling each other because the matrices are orthonormal.
Otherwise, we use the strict convexity of $\mathcal{L}^l$ with respect to $B^l$, which implies that $B_*^l$ is uniquely defined. Then using the orthonormality of $W^l$, it means that the norm of $(W^L \cdots  W^{l+1})^\top B^L (W^L \cdots W^{l+1})$ is the norm of $B^L$, therefore we have $\mathcal{L}^L(B^L) = \mathcal{L}^l \left((W^L \cdots  W^{l+1})^\top B^L (W^L \cdots W^{l+1})\right)$.
Therefore if we denote $B_*^L$ the minimum of $\mathcal{L}^L$, this equality implied that the minimum of $\mathcal{L}^l$ is reached for
$B_*^l = (W^L \cdots  W^{l+1})^\top B_*^L (W^L \cdots W^{l+1}) $.
\end{proof}

\subsection{Proof of theorem \ref{th:low_dim}}

Our goal is to show that the local-SSL gradients are closer to the BP-SSL gradient when we use the direct feedback.
To avoid ambiguity in the context of the following proof we use the matrix $B^l$ to refer to matrix of the local-SSL $ \mathcal{L}^l(B^l)$ and $D^l$ to refer to the local-SSL DFB $ \mathcal{L}^l(D^l)$. For all layers $l < L$, $D^l$ is typically a rectangular matrix while $B^l$ is a square matrix.
Given our notation the loss at the top layer is identical in both, and it is equal to the loss used with BP-SSL, so we have by definition: $\BPL_* = \mathcal{L}_*^L = \mathcal{L}_{*,fb}^L $. 
And we seek to show that:

\begin{equation}
        \norm{ \frac{\partial \mathcal{L}_{*}^l}{\partial W^l} - \frac{ \partial \BPL_*}{\partial W^l} }_F^2  \geq \norm{\frac{\partial \mathcal{L}_{\rm *, fb}^l}{\partial W^l} - \frac{ \partial \BPL_*}{\partial W^l}}_F^2
\end{equation}

\begin{proof}
    % We assume a loss of type 1 with rectified linear non-linearity.
    % First we will show that if any of the loss is saturated there all the losses are saturated so we can focus on the study the problem in the linear range of the function $f(\cdot)$.
    % First we will first show that the losses $\mathcal{L}_*^l$ is saturated if and only if $\mathcal{L}_*^L$ is saturated.

     We denote $R^l = W^L W^{L-1} \cdots W^{l+1}$. The dimensionality of each matrix $W^l$ is $n^{l+1} \times n^{l}$ and $n^{l+1} < n^l$. The reduced dimensionality forces the last layer activity to compute scores in a low-dimensional space. We show that the null-space of $R^l$ leads to difference between layerwise and end-to-end gradient.
    
    We do singular value decomposition of $R^l = U SV^T$. Here $S$ is a $n^L$-by-$n^{l+1}$ matrix with first $n^L$ dimensions being identity matrix $I_{n^L}$ and zeros elsewhere. Then, we could rewrite it also as $US_lV_l^T$ with $S_l$ being $I_{n^L}$ and $V_l$ being the first $n^L$ columns of $V$. So, we have
    $R^l = U V_l^T$. In addition, we denote $V_\emptyset$ the remaining columns of $V$.

    We do a change of variable $B^l = V \hat{B} V^T$ and examine the layerwise loss with a single input: (for simplicity, we denote $\Delta z^l = z_\text{neg}^{l} - z_\text{pos}^{l}$)
    \begin{align*}
        \mathcal{L}_*^l & = \operatorname{min}_{\hat{B}} (\Delta z^l)^T V\hat{B}V^T c^{l}  + \lambda ||\hat{B}||_F^2
    \end{align*}
    We could rewrite $V$ as $[V_l \  V_\emptyset]$ and $\hat{B}$ as 
    $\begin{bmatrix}
    B_1 & B_2\\
    B_3 & B_4
    \end{bmatrix}$ with $B_1, B_2$ having $n^L$ rows and $B_1, B_3$ having $n^L$ columns. Plug it into the loss:
    \begin{align*}
        \mathcal{L}_*^l & = \operatorname{min}_{B_1..B_4} (\Delta z^l)^T [V_l \  V_\emptyset]
        \begin{bmatrix}
        B_1 & B_2\\
        B_3 & B_4
        \end{bmatrix}
        \begin{bmatrix}
        V_l^T\\
        V_\emptyset^T
        \end{bmatrix} c^{l}  + \lambda \sum_{j=1}^4||B_j||_F^2\\
        & = \operatorname{min}_{B_1..B_4} (\Delta z^l)^T  (V_lB_1V_l^T + V_lB_2V_\emptyset^T + V_\emptyset B_3V_l^T + V_\emptyset B_4 V_\emptyset^T)c^{l}  + \lambda \sum_{j=1}^4||B_j||_F^2
    \end{align*}

    If we isolate all the terms depending on $B_1$, we show that this part is equivalent to BP loss $\BPL$. By using a change of variable $\hat{W} = UB_1U^T$:
    \begin{align*}
        \mathcal{L}_{*, B_1}^l &= \operatorname{min}_{B_1}\sum_\mu (\Delta z^l)^T  V_l B_1V_l ^T c^{l}  + \lambda ||B_1||_F^2 \\
        & = \operatorname{min}_{\hat{W}}(\Delta z^l)^T  V_l U^T \hat{W } U V_l ^T c^{l}  + \lambda ||\hat{W}||_F^2 \\
        & = \operatorname{min}_{\hat{W}}(\Delta z^l)^T  R_l^T \hat{W } R^l c^{l}  + \lambda ||\hat{W}||_F^2 \\
        & =\operatorname{min}_{\hat{W}}(\Delta z^L)^T \hat{W } c^{L}  + \lambda ||\hat{W}||_F^2 = \mathcal{L}_{*}^L = \BPL_*
    \end{align*}

    Therefore, the difference between $\mathcal{L}_*^l$ and $\BPL_*$ are the $B_2, B_3, B_4$ terms:
    \begin{align*}
        \mathcal{L}_*^l  - \BPL_* = \operatorname{min}_{B_2..B_4} (\Delta z^l)^T  ( V_lB_2V_\emptyset^T + V_\emptyset B_3V_l^T + V_\emptyset B_4 V_\emptyset^T)c^{l}  + \lambda \sum_{j=2}^4||B_j||_F^2
    \end{align*}

    Similarly, we apply the same methods to identify the difference between $\mathcal{L}_{*, fb}^l$ and $\BPL_*$. To better differentiate the $B^l$ in $\mathcal{L}^l$ and $\mathcal{L}_{fb}^l$, we use $D^l$ to denote the rectangular matrix in $\mathcal{L}_{fb}^l$. The definition for $\mathcal{L}_{*, fb}^l$ is then:
    \begin{align*}
        \mathcal{L}_{*, fb}^l = \operatorname{min}_{\hat{D}} (\Delta z^l)^T  D^l c^{L}  + \lambda ||D^l||_F^2 
    \end{align*}
    
    We again perform the change of variable $D^l = V\hat{D}$:

    \begin{align*}
        \mathcal{L}_{*, fb}^l & = \operatorname{min}_{\hat{D}} (\Delta z^l)^T  V\hat{D} c^{L}  + \lambda ||\hat{D}||_F^2 \\
        & = \operatorname{min}_{D_1, D_2}\sum_\mu (\Delta z^l)^T  [V_l \  V_\emptyset]
        \begin{bmatrix}
        D_1 \\
        D_2
        \end{bmatrix} c^{L}  + \lambda \sum_{j=1}^2||D_j||_F^2\\
        & = \operatorname{min}_{D_1, D_2} (\Delta z^l)^T  (V_l D_1 + V_\emptyset D_2 )c^{L}  + \lambda \sum_{j=1}^2||D_j||_F^2 \\
        & = \BPL_* + \operatorname{min}_{D_2}(\Delta z^l)^T V_\emptyset D_2 c^{ L}  + \lambda ||D_2||_F^2
    \end{align*}

    Here, the equality of $\BPL_*  = \operatorname{min}_{D_1} (\Delta z^l)^T V_\emptyset D_1 c^{L}  + \lambda ||D_1||_F^2$ can be obtained from following the same proof of $\mathcal{L}_{*, B_1}^l = \BPL_*$ and a change of variable $D_1 = U^T \hat{\mathcal{W}}$. 
    
    For the other term, we do change of variable $ D_2 = \tilde{D_3} U^T$, then:
    \begin{align*}
        \mathcal{L}_{*, fb}^l - \mathcal{L}_{*}^L &= \operatorname{min}_{D_3} (\Delta z^l)^T V_\emptyset D_3 U^T R^l c^l  + \lambda ||D_3||_F^2 \\
        &= \operatorname{min}_{D_3} (\Delta z^l)^T V_\emptyset D_3 V_l^T c^{l}  + \lambda ||D_3||_F^2 
    \end{align*}

    Notice these are the same as parts with $B_3$ in layerwise difference $\mathcal{L}_*^l - \BPL_*$. So there is the additional term with $B_2$ and $B_4$:

    \begin{align*}
        \mathcal{L}_*^l  - \BPL_* = \mathcal{L}_{*, fb}^l - \BPL_* + \operatorname{min}_{B_2, B_4} (\Delta z^l)^T (V_lB_2V_\emptyset^T + V_\emptyset B_4V_\emptyset^T) c^{l}  + \lambda (||B_2||_F^2 + ||B_4||_F^2)
    \end{align*}

    This additional deviation in loss leads to additional deviation in gradient. 

    By definition of $B_{*,2}$ and  $B_{*,4}$ of when computing the difference of gradient with respect to $W^l$, we obtain the two gradients deviations for $\mathcal{L}_{*}^l$ and $\mathcal{L}_{*, fb}^l$ respectively: (for simplicity, we denote $\Delta z^{\mu, l-1} = z_\text{neg}^{\mu, l-1} - z_\text{pos}^{\mu, l-1})$
    \begin{align*}
        \frac{\partial \mathcal{L}_{*, fb}^l}{\partial W^l} - \frac{\partial \BPL_*}{\partial W^l} &=  V_\emptyset D_{*, 3}V_l^T c^{l} (\Delta z^{l-1})^T \\
        \frac{\partial \mathcal{L}_{*}^l}{\partial W^l} - \frac{\partial \BPL_*}{\partial W^l} 
         & = (V_lB_{*, 2} V_\emptyset^T +  V_\emptyset B_{*, 4} V_\emptyset^T) c^{l} (\Delta z^{l-1})^T + \frac{\partial \mathcal{L}_{*, fb}^l}{\partial W^l} - \frac{\partial \BPL_*}{\partial W^l}
    \end{align*}
In the following proof, we want to show that the two extra terms make the gradient deviate more instead of cancelling the deviation from $\mathcal{L}_{*, fb}^l$. 

To begin with, we prove that the first extra term $V_lB_{*, 2} V_\emptyset^T c^l (\Delta z^{l-1})^T$ is orthogonal to the rest. Note that $\frac{\partial \mathcal{L}_{*, fb}^l}{\partial W^l} - \frac{ \partial \BPL_*}{\partial W^l}$ is in the form of $V_\emptyset$ multiplied with other matrices, so each column vector of $\frac{\partial \mathcal{L}_{*, fb}^l}{\partial W^l} - \frac{ \partial \BPL_*}{\partial W^l}$ is in the column space of $V_\emptyset$ (i.e., the subspace spanned by the column vectors of $V_\emptyset$. The statement is also true for the second extra term: $V_\emptyset B_{*, 4}V_\emptyset^T c^{l} (\Delta z^{l-1})^T$. In contrast the first extra term $V_lB_{*, 2} V_\emptyset^T c^l (\Delta z^{l-1})^T$ is in the column space of $V_l$, which is orthogonal to the column space of $V_\emptyset$. Then:
\begin{align*}
        \norm{\frac{\partial \mathcal{L}_{*}^l}{\partial W^l} - \frac{ \partial\BPL_*}{\partial W^l}}_F^2 &= \norm{\frac{\partial \mathcal{L}_{*, fb}^l}{\partial W^l} - \frac{ \partial\BPL_*}{\partial W^l} +  V_\emptyset B_{*, 4}V_\emptyset^T c^{l} (\Delta z^{l-1})^T}_F^2 + \norm{ V_l B_{*, 2}V_\emptyset^T c^{l} (\Delta z^{l-1})^T}_F^2\\ &\geq \norm{\frac{\partial \mathcal{L}_{*, fb}^l}{\partial W^l} - \frac{ \partial\BPL_*}{\partial W^l} + V_\emptyset B_{*, 4}V_\emptyset^T c^{l} (\Delta z^{l-1})^T}_F^2
    \end{align*}

Next, we show that $V_\emptyset B_{*, 4}V_\emptyset^T c^{l} (\Delta z^{l-1})^T$ does not cancel the deviation from $\frac{\partial \mathcal{L}_{*, fb}^l}{\partial W^l} - \frac{ \partial\BPL_*}{\partial W^l}$. To do so, we first compute analytically the expression for $B_{*, 4}$. Since function $V_\emptyset B_{ 4}V_\emptyset^T c^{l} (\Delta z^{l-1})^T + \lambda\norm{B_4}_F^2$ is strictly convex, globally minimal $B_{*, 4}$ is reached when the expression is at critical point (zero derivative):
\begin{align*}
     &\frac{\partial}{\partial B_4} \big(\Delta (z^{l})^T  V_\emptyset B_{ 4}V_\emptyset^T c^{l}  + \lambda\norm{B_4}_F^2\big) = 0 \text{  when evaluated at } B_{*, 4}\\
    \Leftrightarrow  & \ \  0 = V_\emptyset^T \Delta z^{l} (c^{l})^T V_\emptyset + 2 \lambda B_{*, 4} \\
    \Leftrightarrow & \ \   B_{*, 4} = -\frac{1}{2\lambda} V_\emptyset^T \Delta z^{l} (c^{l})^T V_\emptyset    
\end{align*}

Following the same process, we can obtain $D_{*, 3} = -\frac{1}{2\lambda} V_\emptyset^T \Delta z^{l} (c^{l})^T V_l$.

Substituting the analytical expression of $B_{*, 4}$ and $D_{*,3}$ into deviations of gradients, we could derive:
\begin{align*}
        \norm{\frac{\partial \mathcal{L}_{*}^l}{\partial W^l} - \frac{ \partial\mathcal{L}_{ *}^L}{\partial W^l}}_F^2 & \ge \norm{\frac{\partial \mathcal{L}_{*, fb}^l}{\partial W^l} - \frac{ \partial\mathcal{L}_{ *}^L}{\partial W^l} + V_\emptyset B_{*, 4}V_\emptyset^T c^{l} (\Delta z^{l-1})^T}_F^2 \\
        & = \norm{V_\emptyset D_{*, 3}V_l^T c^{l} (\Delta z^{l-1})^T + V_\emptyset B_{*, 4}V_\emptyset^T c^{l} (\Delta z^{l-1})^T}_F^2\\
        & = \norm{-\frac{1}{2\lambda }  V_\emptyset V_\emptyset^T \Delta z^{l} (c^{l})^T ( V_\emptyset V_\emptyset^T +  V_l V_l^T) c^{l} (\Delta z^{l-1})^T}_F^2\\
        & = \norm{-\frac{1}{2\lambda }  V_\emptyset V_\emptyset^T \Delta z^{l} (\norm{V_\emptyset^T c^l}_2^2 + \norm{V_l^T c^l}_2^2) (\Delta z^{l-1})^T}_F^2\\
        & \ge \norm{-\frac{1}{2\lambda }  V_\emptyset V_\emptyset^T \Delta z^{l} (\norm{V_l^T c^l}_2^2) (\Delta z^{l-1})^T}_F^2\\
        &  = \norm{-\frac{1}{2\lambda }  V_\emptyset V_\emptyset^T \Delta z^{l} (c^{l})^T (V_l V_l^T) c^{l} (\Delta z^{l-1})^T}_F^2 = \norm{\frac{\partial \mathcal{L}_{*, fb}^l}{\partial W^l} - \frac{ \partial\mathcal{L}_{ *}^L}{\partial W^l}}_F^2
    \end{align*}

\end{proof}

\subsection{Proof of proposition \ref{prop:spatial_dependence}}
Consider the layerwise loss function with 2D spatial dependence and patch size 1, the loss can be expressed in the form ${\cal L}^l = f(\flatten(z^l)^T B^l \flatten (c^l))$. Hence, it results in a large feedback matrix $B^l$ enabling all-to-all connectivity across the feature maps. Here we omit the pos/neg subscript for simplicity. The proof can be applied for both positive and negative inputs, and the proposition is true when both positive and negative components have exact gradient alignment. 

The true BP-SSL gradient is written as: 
\begin{align*}
    \partial {\bf L} / \partial z^{l} = f'(\flatten({z^L})^T B^L \flatten(c^L)) \operatorname{Conv}^T(W^{l+1}, \cdots \operatorname{Conv}^T(W^L, B^L \flatten(W^L * \cdots * W^{l+1} * c^l)))
\end{align*}
where $\operatorname{Conv}^T(W, \cdot) $ is the transpose convolution operation using kernel $W$. We next show that the composition of matrix multiplications and convolution operators can be computed as one large linear operator $B^l$. Mathematically, applying a convolutional kernel $W^l$ could be equivalently represented as multiplying a linear matrix, which has a weight-shared block structure, to the flattened feature map. Denoting the linear matrix as $W_*^l$, we could write BP-SSL gradient as:
\begin{align*}
    \partial {\bf L} / \partial z^{l} = f'(\flatten({z^L})^T B^L \flatten(c^L))  ({W_*^{l+1}}^T \cdots  {W_*^{L}}^T) B^L (W_*^L \cdots W_*^{l+1}) \flatten(c^l)
\end{align*}
We can thereby construct $B^l = {W_*^{l+1}}^T \cdots  {W_*^{L}}^T B^L  W_*^{L} \cdots  W_*^{l+1}$, which enables the gradients of local-SSL ($\partial {\cal L}^l/\partial z^l$) to be equivalent to the gradients of BP-SSL: (for readability, we use $\tilde{z}$ to represent $\flatten(z)$ below)
\begin{align*}
    \partial {\cal L}^l / \partial z^{l} &= f'({\tilde{z^l}}^T B^l \tilde{c^l})  B^l \tilde{c^l} \\
     & = f'({\tilde{z^l}}^T ({W_*^{l+1}}^T \cdots  {W_*^{L}}^T) B^L (W_*^L \cdots W_*^{l+1}) \tilde{c^l})  ({W_*^{l+1}}^T \cdots  {W_*^{L}}^T) B^L (W_*^L \cdots W_*^{l+1}) \tilde{c^l}\\
    & = f'({\tilde{z^l}}^T  B^L  \tilde{c^l})  ({W_*^{l+1}}^T \cdots  {W_*^{L}}^T) B^L (W_*^L \cdots W_*^{l+1}) \tilde{c^l}\\
     & = \partial {\bf L}/\partial z^l
\end{align*}
However, when there is no spatial dependence: $\partial {\cal L}^l / \partial z_{m, n}^{l} = f'(\pool({z^l})^T B^l \pool(c^l))  B^l \pool(c^l)$ for each feature vector $z_{m, n}^l$ at spatial location $m, n$ in layer $l$. This means that all spatial locations receive the same gradient. In contrast, since $\{{W_*^{l}}^T\}_{l=1}^L$ are not necessarily spatially invariant (for convolutional operation with kernel $ >1$, adjacent spatial locations have different weights), gradients of BP-SSL $\partial {\bf L} / \partial z^{l}$ is not necessarily spatially invariant. As a result, in the case of no spatial dependence, there doesn't necessarily exist a $B^l$ such that local-SSL equals BP-SSL gradients.

\section{Training details}
\label{appendix:train}

\paragraph{Simulation for deep linear networks}

For Figure \ref{fig_thm1} and Figure \ref{fig_fb}A, we simulate a 6 layer deep linear network. For Figure \ref{fig_thm1}, each layer has 128 neurons. For Figure \ref{fig_fb}A, the layer width is reduced by half in each layer (from 128 to 4). We took 1024 MNIST samples and split them into 32 batches. For each batched input, images go through only simple data augmentations of cropping 14x14 patches from the image. Patches from the same image are used as positive samples. We use type 2 loss with softplus $f(x) = \log(1+\exp(-x))$ because it is smooth, but the result would hold for other decreasing convex $f$. The regularization coefficient is $\lambda = 0.01$. To compute $B_*^l$, we first use one forward pass to obtain $z^l$ and $c^l$. Then, we use the LBFGS optimizer to find the minimum $B_*^l$. Afterwards, we compute the gradients using $B_*^l$.

For linear CNN (Figure \ref{fig_fb}B), we use 4 layers of convolutional networks, each layer with 32 channels. The kernel sizes and strides are 2. MNIST inputs are randomly cropped into 16x16 patches. Patches from the same image are used as positive samples. We use the same type 2 loss with softplus, with regularization $\lambda = 0.02$. When we train the model with spatial dependence,  $z^1, c^1$ features in the local 2x2 patches share the same $B^1$ in the first layer, but no features share $B^l$ in subsequent layers (full 2D spatial dependence). This is effectively implemented by performing local $2 \times 2$ local average pooling for $z^1$ and $c^1$, and flattening $z^l, c^l$ in other layers.

\paragraph{Simulation for optimum similarity with BP update (Figure \ref{fig_fb}B)}

We simulate a 6 layer ReLU MLP with 512 neurons in each layer. The training dataset in MNIST is used to train the network. For each batched input, images go through only simple data augmentations of cropping 14x14 patches from the image. We use again type 2 loss with softplus $f(x) = \log(1+\exp(-x))$. The batch size is 32, and we use Adam optimizer with a learning rate is $5 \times 10^{-5}$. Each model is trained for 20 epochs with different versions of local-SSL algorithms. Following training, each model use the test dataset of MNIST to evaluate cosine similarity of its gradient updates to the BP updates. For the theoretical optimum, we just change the update rule of $B_\circ^l$ to the gradient descent of loss $\norm{
    \frac{\partial \mathcal{L}_{fb}^l}{\partial z^{l}} (B_\circ^l) - \frac{\partial \mathcal{L}_*^L}{\partial z^{l}}
   }_F^2$ as defined in \ref{def:optimal_fb}.

\paragraph{For Table \ref{tab: clapp_compare}, on CIFAR10, STL10, and TinyImageNet}

 We follow the training setup and procedure of \citet{illing_2021}. We use the STL-10 dataset \citep{coates2011analysis}, which is a dataset designed for unsupervised learning. It contains 100,000 unlabeled images. It also contains 500 labeled training images and 800 test images for each class. We first train the neural network model (Table \ref{tab: arch}) on the unlabeled images using the self-supervised losses. Then, we freeze the pretrained neural network, and then train a linear classifier using the representations of labeled training images. The decoding accuracy performance is computed by evaluating the decoder on representations of the test images.

\begin{table}[h]
  \caption{Network Architecture}
  \label{tab: arch}
  \centering
  \begin{tabular}{cc}
    \toprule
    \# of trainable layer & layer type\\
    \midrule
    1 & 3x3 conv128, ReLU\\
    \midrule
    2 & 3x3 conv256, ReLU\\
     & 2x2 MaxPool\\
    \midrule
    3 & 3x3 conv256, ReLU\\
    \midrule
    4 & 3x3 conv512, ReLU\\
     & 2x2 MaxPool\\
     \midrule
    5 & 3x3 conv1024, ReLU\\
     & 2x2 MaxPool\\
     \midrule
    6 & 3x3 conv1024, ReLU\\
     & 2x2 MaxPool\\
    \bottomrule
  \end{tabular}
\end{table}

To create positive or negative samples, each image undergoes the data augmentations as in SimCLR \citep{chen2020simclr}. Specifically, we use the torchvision.transforms in PyTorch to perform the following augmentations:  RandomResizeCrop, RandomHorizontalFlip, random ColorJitter, RandomGrayScale, random GaussianBlur.

For introducing 2D spatial dependence, we let features within the local $n \times n$ (size in Table \ref{tab: pool}) patches to share $B^l$. This is effectively implemented by performing local average pooling depending on the layer and the image dataset.

\begin{table}[h]
  \caption{Size of local pooling kernel (i.e., patches sharing $B^l$) at different layers (L1-L6) for computing the loss}
  \label{tab: pool}
  \centering
  \begin{tabular}{c|c|cccccc}
    \toprule
    Dataset & Image Size & L1 & L2 & L3 & L4 & L5 & L6\\
    \midrule
    CIFAR10& 32 &4 & 4 & 4 & 2 & 2 & 1\\
    STL10& 96 & 12 & 12 & 12 & 6 & 6 & 3\\
    Tiny-ImageNet& 64 & 8 & 8 & 8 & 4 & 4 & 2\\
    \bottomrule
  \end{tabular}
\end{table}

For BP-CLAPP++ training, we just take the CLAPP++ objective at the last layer, and then perform backpropagation to train earlier layers. For BP-InfoNCE, we use a linear projection head with output dimension 1024 to project $z^L$ and $c^L$. Then, we use the InfoNCE loss as in \citet{chen2020simclr} and backpropagation to train the whole network. To examine the influence of spatial dependence in BP-SSL training, we simulate both the case with spatial dependence and without for $\BPL$. We found no effect of spatial dependence on performance, so we just report the number without spatial dependence.

The model receives 128 images at each batch during training. For each patch of each image, one positive contrastive sample and one negative contrastive sample are selected to compute the loss $\mathcal{L}^l$ and to update the network parameters. We use Adam optimizer with a learning rate $2\cdot10^{-4}$ to train each model for 300 epochs. For STL10, each experiment takes approximately 5 hours on 4 NVIDIA A100 GPUs.

For training the linear classifier, the representation from each layer is first average pooled using the same kernel as during training, and then the representations across all layers are concatenated as input for the linear classifier. We use Adam optimizer with a learning rate $10^{-3}$.

\paragraph{For Table 2, on ImageNet}
We used the 8 layer convolutional networks as in VGG11 \cite{simonyan2015vgg}, but we doubled the network width.  When introducing 2D spatial dependence, we implement local average pooling using the AdaptiveAvgPool2d function of PyTorch, which adapts pooling kernels and strides to a targeted output pooling dimension. The architecture and the output pooling dimension are specified in Table \ref{tab: arch_imgnet}. 

\begin{table}[h]
  \caption{Network Architecture and adaptive local spatial pooling size for ImageNet experiments}
  \label{tab: arch_imgnet}
  \centering
  \begin{tabular}{ccc}
    \toprule
    \# of trainable layer & layer type & pooling dimension\\
    \midrule
    1 & 3x3 conv128, ReLU & 8\\
    & 2x2 MaxPool &\\
    \midrule
    2 & 3x3 conv256, ReLU & 4\\
     & 2x2 MaxPool & \\
    \midrule
    3 & 3x3 conv512, ReLU & 4\\
    \midrule
    4 & 3x3 conv512, ReLU & 4\\
     & 2x2 MaxPool & \\
     \midrule
    5 & 3x3 conv1024, ReLU & 2\\
     \midrule
    6 & 3x3 conv1024, ReLU & 2\\
     & 2x2 MaxPool & \\
     \midrule
     7 & 3x3 conv1024, ReLU & 2\\
     \midrule
    8 & 3x3 conv1024, ReLU & 2\\
     & 2x2 MaxPool & \\
    \bottomrule
  \end{tabular}
\end{table}

We use the same data augmentations as other image classification experiments, except that the image size is cropped and resized to 224. We use a batch size of 256 and train the model for 100 epochs. Following \citet{chen2020simclr}, we use the LARS optimizer with a learning rate of 0.3 and a weight decay of $10^{-6}$. We also use cosine-annealed learning rate scheduler with linear warmup in the first 10 epochs. Each experiment runs approximately 12 hours on 4 NVIDIA H100 GPUs. 

For training the linear classifier, the representation from each layer is first average pooled along the spatial dimensions before being concatenated together. We use Adam optimizer with a learning rate $10^{-3}$.

\paragraph{Simulation for Figure \ref{fig_cnn}C}:
We repeat the experiments on STL10, except that we remove spatial dependence in the last layer for `CLAPP++' and `CLAPP++DFB'. In this way, they use the same last layer objective ($\BPL$) as model `CLAPP++ (no spatial dependence)', so the comparison with BP-SSL gradients are fair across these learning rules. The classification performance on STL10 does not change if we remove the spatial dependence of the last layer. CLAPP++ reaches 78.63\% accuracy, and CLAPP++DFB reaches 79.01\% accuracy. After simulation, we freeze the encoder at 100 epochs (or other number of epochs in Figure \ref{fig:grad_by_ep}) and evaluate the gradient alignment to BP-CLAPP++ on the held out test dataset.

\section{Training self-supervised objectives with BP-approximation methods}

\label{appendix:pc}

 We compare our local-SSL rules to two existing biologically plausible BP-approximation frameworks: Directed Feedback Alignment (DFA) \citep{nokland2016dfa} and Predictive Coding (PC) \citep{whittington2017predictivecoding}. More specifically, we use these methods as approximations of BP-CLAPP where the CLAPP objective is applied in the final output layer.

\paragraph{Self-supervised Directed Feedback Alignment} In DFA, the partial derivative of the loss with respect to the final layer activity $\delta^L = \partial \BPL/\partial z^L$ is directly projected to each intermediate layer through a random fixed matrix: $\delta^l = F^l \delta^L$. The feedforward weight is then updated with: $\Delta W_{ji}^{l} = \delta_j^{l} \rho'(a_j^{l}) z_i^{l-1}$. When a shallow network is trained using DFA on a simple task like MNIST, the network evolves such that $\delta^{l}$ in intermediate layers would approximate the true partial derivative $\partial \BPL / \partial z^{l}$.

Applying this method to optimize the final layer contrastive predictive loss $\BPL$ of CLAPP, we get the update rule for feedforward weights in the intermediate layers:

\begin{equation}
    \label{eq:clapp_dfa}
    \Delta W_{ji}^{l} = \gamma^L \cdot (F^l B^{L} z^{L})_j \cdot \rho'(a_j^{l}) z_i^{l-1} 
\end{equation}

Such a method based on feedback alignment is known to be unscalable to deeper networks and more complex image classification tasks  \citep{bartunov2018assessing}. In Table \ref{tab: clapp_compare}, we also show that it performs much inferior to the CLAPP model. 

The method is closely related to our proposed feedback scheme. In the CLAPP++DFB learning rule, $B^l$ is used to replace the combined term $F^lB^L$ in the DFA method. However, the top-down $B^l$ in CLAPP++DFB is trained with a Hebbian learning rule, whereas the top-down connection $F^l$ in DFA is a random fixed matrix. The broadcasting factor $\gamma^l$ in CLAPP++DFB also has dependence on layer activities.

\paragraph{Self-supervised Predictive Coding}

In short, we use the JAX package for predictive coding implemented in \citet{Pinchetti2025benchmark} (\url{https://github.com/liukidar/pcx}) to optimize the self-supervised objective. 

This method takes a different approach by modeling each layer's activity $h^l$ as a Gaussian distribution conditioned on the previous layer's activity and layer parameters: $P(h^l) \sim  \mathcal{N}(\mu^l, \Sigma^l)$ where $\mu^l =f(\theta^l, h^{l-1}) = W^l \rho(h^{l-1})$ for a feedforward layer, and $\Sigma^l$ is usually set as identity matrix. In predictive coding, activity $h^l$ and parameters $\theta^l$ are optimized to minimize the following energy function:

% In supervised learning, the network maximizes the likelihood of intermediate layer activity conditioned on both the input $x$ and target $y$: $P(h^1 \dots h^{L-1}|h^0 = x, h^L=y)$. If the output layer is modeled by any arbitrary distribution instead of the Gaussian distribution \citep{pinchetti22_arbitrary}, the likelihood maximization is equivalent to minimizing the energy function $\mathcal{F}$:

\begin{equation}
    \label{eq:pca_learning}
    \mathcal{F}(h, \theta) = \sum_{l=1}^{L-1} \frac{1}{2}(h^l - \mu^l)^2 + \mathcal{L}(\mu^L, y)
\end{equation}

% Here $\mathcal{L}$ could be any arbitrary loss function defined at the final layer $L$. To apply predictive coding to the contrastive predictive loss, we could simply set $L$ to the contrastive predictive loss

% \begin{equation}
%     \label{eq:pca_clapp_loss}
%     \mathcal{L}(\mu^L, y) = \mathcal{L}^{t, L}_{cp}(\mu^l) = \max(0, 1 - y^t \cdot u_{t-\delta t}^{t, L}), \ \ u_{t-\delta t}^{t, L} = {\rho(\mu^{t, l}})^T W^{\text{pred}, l} \rho(\mu^{t -\delta t, l})
% \end{equation}

Here $\mathcal{L}$ could be the loss function defined at the final layer $L$. The classic predictive coding method \citep{whittington2017predictivecoding} optimizes the energy $\mathcal{F}$ by first updating the activity $h^l$ to an equilibrium and then updating the parameters $\theta^l$. The feedback information is propagated down from the top layer when the activities $h^l$ are updated using the gradient descent of $\mathcal{F}$:

\begin{equation}
    \label{eq:pca_h_update}
    \Delta h^l = \eta_h [-\epsilon^l + \rho'(h^l) \odot ({W^{l+1}}^T \epsilon^{l+1})]
\end{equation}

where $\epsilon^l = h^l - \mu^l$ for $l<L$ and $\epsilon^L = -\partial \mathcal{L}/ \partial\mu^L$. The transpose of the feedforward weight ${W^{l+1}}^T$ propagates down the information $\epsilon^L$ from the final layer. After updating the activity to equilibrium $h_*^l$, the parameters are updated using the gradient descent, which has the Hebbian format:

\begin{equation}
    \label{eq:pca_w_update}
    \Delta W^l = \eta_\theta \cdot \rho(h^{l-1}) \epsilon^l
\end{equation}

We apply this method to optimize the self-supervised CLAPP objective at the final layer by replacing $\mathcal{L}(\mu^L, y) $ with $\BPL(\rho(\mu^L))$. To optimize the energy during training, we follow the incremental predictive coding (iPC) method used in \citet{Salvatori2024ipc} to update the activity $\{h^l\}_{l=1}^{L-1}$ and parameters $\{W^l\}_{l=1}^{L}$ iteratively. This method has been shown to achieve better performance and more stable training. It is also more biologically plausible than the classic predictive coding \citep{whittington2017predictivecoding} or 
 equilibrium propagation \citep{scellier2017equilibrium} because it does not require additional signals to control separate phases of updating activities and updating parameters.

 \begin{table}[bh]
  \caption{
  Image classification performance using backpropagation (BP) and predictive coding (PC) to optimize different losses. }
  \label{tab: pc_all}
  \centering
  \begin{tabular}{ccccc}
    \toprule
     & \multicolumn{2}{c}{CIFAR-10} & \multicolumn{2}{c}{STL-10}                   \\
    \cmidrule(r){2-3} \cmidrule(r){4-5} 
    Loss     & BP     & PC & BP & PC \\
    \midrule
    Cross-entropy (supervised) & 72.41  & 71.92 & - &  -   \\
    CLAPP objective  (self-supervised)   &  65.68 & 45.62 &  78.36 & 36.75   \\
    LPL objective  (self-supervised)  & 73.65 & 47.68 & 70.29 & 51.21  \\
    \bottomrule
  \end{tabular}
\end{table}

\begin{table}[ht]
  \caption{
  Hyperparameters for different losses in predictive coding. $T$ is the number of updates for $h$ and $w$ for each input. $\eta_h$ is the learning rate for updating $h$. These two hyperparameters are specific to predictive coding and are thereby tuned. Other hyperparameters are kept the same as the ones used in backpropagation training.}
  \label{tab: hyper_pc_loss}
  \centering
  \begin{tabular}{c|ccc}
    \toprule
    Loss & PC-Supervised & PC-CLAPP & PC-LPL\\
    \midrule
    T & \multicolumn{3}{c}{\{1, 9, 17\}}\\
    $\eta_h$ & \multicolumn{3}{c}{\{0.01, 0.1, 0.5 \}}\\
    \midrule
    Batch size & 128 & 32 & 1024\\
    $\eta_w$ & $5\cdot10^{-5}$ & $2\cdot10^{-4}$ & $10^{-3}$\\
    Epochs & 50 & 300 & 800\\
    Weight decay & $10^{-4}$ & 0 & $10^{-4}$\\
    Optimizer &  \multicolumn{3}{c}{Adam} \\
    \bottomrule
  \end{tabular}
\end{table}

 After performing a hyperparameter search and selecting the best model, we found that predictive coding fails to learn useful representations. We find that predictive coding only performs close to backpropagation on supervised loss, whereas the method fails to optimize the self-supervised contrastive objectives used by CLAPP (Table \ref{tab: pc_all}). To further show that the result is not specific to contrastive objectives like CLAPP, we also implement predictive coding to optimize the non-contrastive objectives used by LPL \citep{halvagal2023lpl}, where we found a similar result (Table \ref{tab: pc_all}). Therefore, the predictive coding method could not be easily extended to self-supervised loss.

To update $h^l$, we follow the code in \citet{Salvatori2024ipc}, and use the SGD optimizer with a momentum of $0.5$. The learning rate $\eta_h$ and optimization steps $T$ are searched according to Table \ref{tab: hyper_pc_loss}. In supervised training, we update $W^l$ using the Adam optimizer with a learning rate $5\cdot10^{-5}$ and weight decay $10^{-4}$. We use a batch size of 128 and train the model for 50 epochs. The test dataset is evaluated at the end of each epoch, and the best accuracy is reported. In self-supervised training, we use the same batch size, epoch number, learning rate, and optimizer for $W^l$ as in CLAPP \citep{illing_2021} and LPL \citep{halvagal2023lpl}. Detailed choices are specified in Table \ref{tab: hyper_pc_loss}:

\section{Additional Comparison with BP-SSL on Local-SSL}
\label{appendix:extra}
We analyze the effect of imposing constraints on $B^l$ for theorem \ref{th1}. We constrain the ranks of $B^l$ by parametrizing $B^l = (F_1^l)^TF_2^l$ where $F_1^l$ and $F_2^l$ are trainable weight matrices, both with $n^l$ columns and $r$ rows ($n^l$ is number of neurons in layer $l$, and $r<n^l$). Then, we repeat the same simulation for theorem \ref{th1} (green line in Figure \ref{fig_thm1}). As the rank decreases, $B^l$ is more constrained, and the alignment between local-SSL and BP-SSL decreases.

\begin{figure}[bh]
  \centering
  \includegraphics[width=0.32\textwidth]{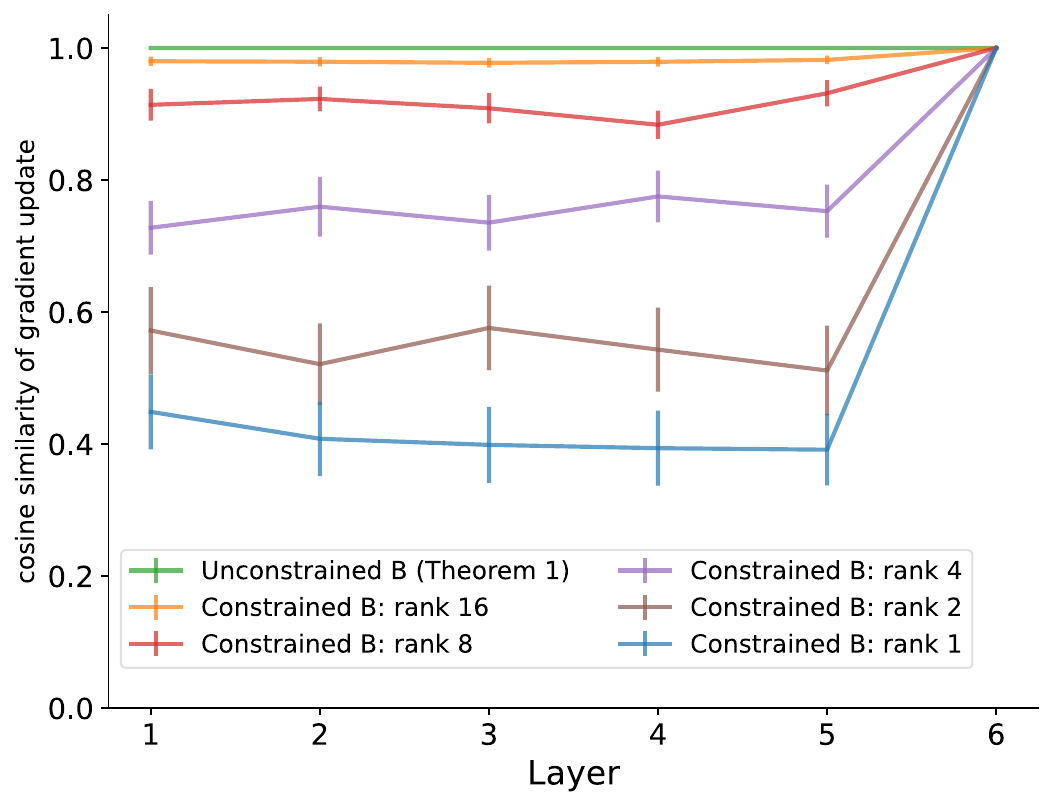}
  \caption{Cosine similarity of weight update of local-SSL and BP-SSL for $B^l$ with different rank. Simulation is done in 6-layer deep linear network with 128 neurons in each layer.}
  \label{fig:rank_constraints}
\end{figure}

\begin{figure}[bh]
  \centering
  \includegraphics[width=0.8\textwidth]{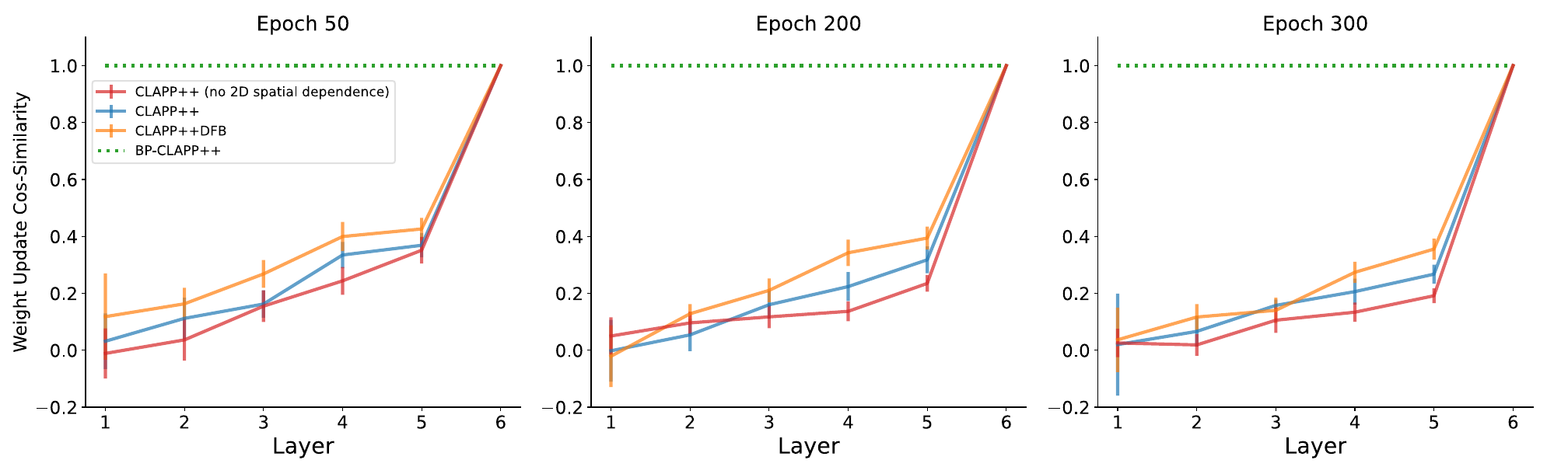}
  \caption{Cosine similarity of weight update of local-SSL and BP-SSL at different stages of the training process on STL-10.}
  \label{fig:grad_by_ep}
\end{figure}

We analyze the gradient alignment at different stages of training in STL10 experiments (Figure \ref{fig:grad_by_ep}). In all stages, introducing direct feedback and 2D spatial dependence improves the gradient alignment.

\end{document}